\documentclass{article}

\PassOptionsToPackage{numbers, compress}{natbib}

 \usepackage[preprint]{neurips_2026}

\usepackage[utf8]{inputenc} %
\usepackage[T1]{fontenc}    %
\usepackage{hyperref}       %
\hypersetup{hidelinks}
\usepackage{url}            %
\usepackage{booktabs}       %
\usepackage{amsfonts}       %
\usepackage{nicefrac}       %
\usepackage{microtype}      %
\usepackage{xcolor}         %
\usepackage{enumitem}       %
\usepackage{graphicx}       %
\usepackage{amsmath}
\usepackage{amssymb}
\usepackage{xcolor}
\usepackage{multirow}       %
\usepackage{wrapfig}        %
\usepackage{soul}
\usepackage[font=small]{caption}
\usepackage{tikz}
\usetikzlibrary{positioning, arrows.meta, calc}

\newcommand{\vjepag}{V-JEPA~2.1}
\newcommand{\egoexo}{Ego-Exo4D}
\newcommand{\egovlp}{EgoVLPv2}
\newcommand{\camformer}{CamFormer}
\newcommand{\viterbinet}{ViterbiPlanNet}
\newcommand{\notraj}{No-Traj}
\newcommand{\retrtraj}{Scorer}
\newcommand{\oracletraj}{Oracle}
\newcommand{\ours}{TrajPilot}
\newcommand{\macc}{\mathrm{mAcc}}
\newcommand{\sr}{\mathrm{SR}}

\title{How You Move Tells What You'll Do: Trajectory-Conditioned Egocentric Prediction}

\author{%
\makebox[\textwidth][c]{%
\begin{tabular*}{0.85\textwidth}{@{\extracolsep{\fill}}cc@{}}
Sejoon Jun\textsuperscript{1,2} &
Hai Nguyen-Truong\textsuperscript{1} \\
\texttt{\small jun.se@northeastern.edu} &
\texttt{\small nguyentruong.h@northeastern.edu} \\[0.5em]
Luigi Seminara\textsuperscript{1,3} &
Lorenzo Torresani\textsuperscript{1} \\
\texttt{\small seminara.l@northeastern.edu} &
\texttt{\small l.torresani@northeastern.edu}
\end{tabular*}
}\\[0.75em]
\makebox[\textwidth][c]{\small \textsuperscript{1}Khoury College of Computer Sciences, Northeastern University, Boston}\\
\makebox[\textwidth][c]{\small \textsuperscript{2}Korea Advanced Institute of Science and Technology, Daejeon}\\
\makebox[\textwidth][c]{\small \textsuperscript{3}Department of Mathematics and Computer Science, University of Catania, Italy}\\\\
\url{https://farsightlab.github.io/TrajPilot/}
}

\begin{document}

\maketitle

\begin{abstract}
Predicting how a person's first-person view will evolve (what action will
follow, what plan completes a task, whether an in-progress shot will
score) is fundamentally under-specified: the same context admits many
plausible futures, and a model trained to minimize prediction error is
forced to hedge or average across them, getting it wrong either way.
Two findings shape our approach. First, the future camera trajectory,
the path the head carves through space, lets the model commit to one of
those futures: it carries the operator's intent in a form fine enough to
determine how an action will unfold, substantially outperforming language
as a conditioning signal. Second, this same intent makes the trajectory
itself partially predictable from the context at hand, enough that
trajectory need not be observed at test time to recover most of the gain.
We instantiate these findings as TrajPilot, a model that predicts
candidate future trajectories from egocentric context and uses them to
pilot action prediction in an action-aligned embedding space where
language shapes the structure but is never used as a conditioning input. TrajPilot beats VLM and structured-planner
baselines on procedural planning across Ego-Exo4D atomic, Ego-Exo4D
Keystep, Ego4D GoalStep, and EgoPER, with the trajectory advantage
widening with horizon (exactly where prior planners collapse) and
holding under RGB-only camera-pose estimation. With the goal masked
at inference, the same model performs goal-free anticipation,
beating VLM baselines on Ego-Exo4D atomic and extending to
EPIC-Kitchens-100 and basketball shot-outcome prediction.
\end{abstract}

\begin{figure}[t]
\centering
\includegraphics[width = \linewidth]{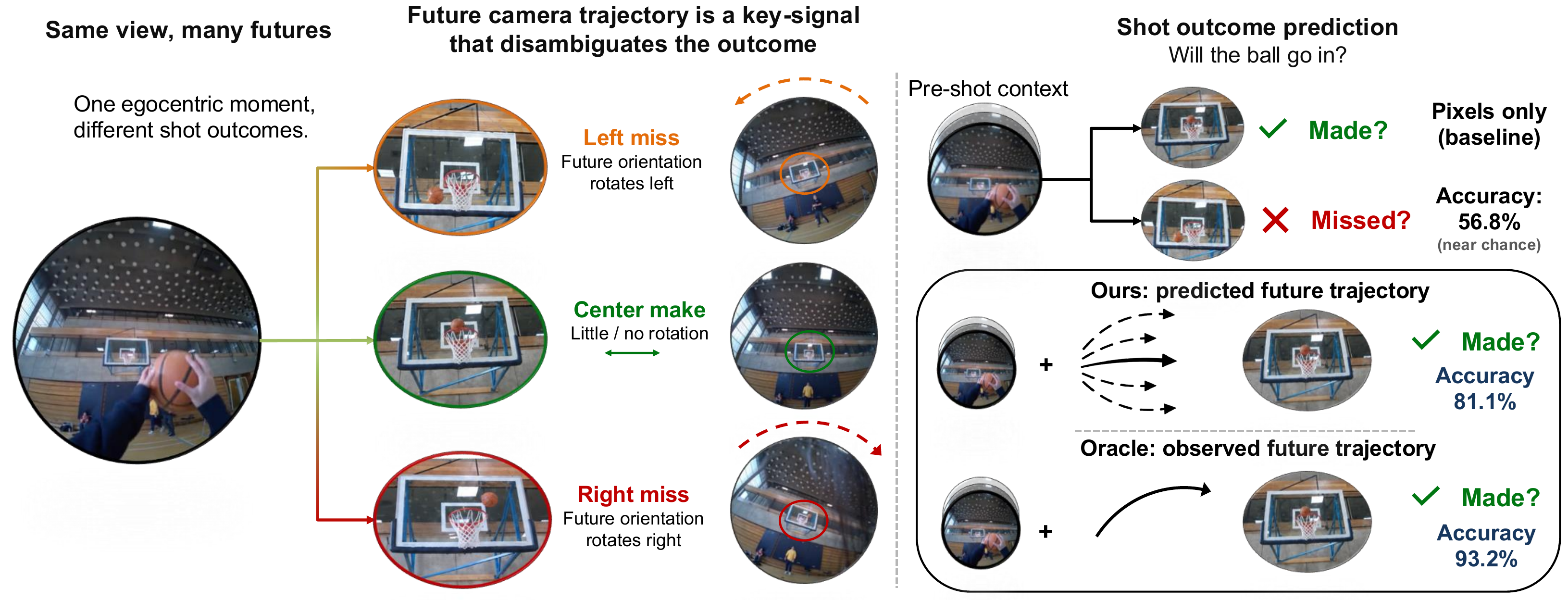}

\caption{A single pre-shot view admits multiple shot outcomes (left).
The future camera trajectory disambiguates them: head rotation is
leftward for a left miss, minimal for a center make, and rightward
for a right miss (middle). On a $74$-shot held-out split, predicting
from pixels alone is near chance ($56.8\%$); given the ground-truth
future trajectory, the model commits to the correct outcome
($93.2\%$, oracle). Future trajectory is unobserved at test time, so
we predict it from the pre-shot context, recovering most of that gain
($81.1\%$). Numbers from Ego-Exo4D basketball (\S\ref{app:basketball}).}
\label{fig:basketball_teaser}
\vspace{-3mm}
\end{figure}

\section{Introduction}
\label{sec:intro}

Imagine an AI assistant that observes a person through a wearable
camera and predicts their near future: whether a basketball shot will
score before release (Figure~\ref{fig:basketball_teaser}), whether a
whisking motion will splatter the bowl,
or whether a wrench-turn will misseat a bracket before damage compounds.
All three are near-future prediction from first-person video, and all
are hard for the same reason: many continuations fit the observed
context, and a model that minimizes prediction error against any single
one hedges or averages, getting it wrong either way.

What signal determines which of these futures will occur? The
conventional answer is language. A description of the wearer's intent,
``the player shoots a jump shot,'' ``the cook whisks the eggs,''
narrows the space of plausible futures, and procedural planners and
action-anticipation models built on this idea have made steady progress
\citep{schema2024,pdpp2023,vedit2025,vlwm2025}. Yet language descriptions
are coarse. There are many ways to ``whisk the eggs,'' and the
difference between a gentle whisk that incorporates them smoothly and an
aggressive one that splatters them out of the bowl is not the
description ``whisk the eggs'' (both attempts share that description)
but the small motor details that determine outcome. The same is true
for the basketball shot and the bracket installation: the description is
shared across success and failure, and what separates them is
\emph{how} the body moves. Whatever signal determines outcome has to be
finer-grained than what language provides.

A surprisingly strong such signal is the \textbf{camera trajectory}:
the path that the wearer's motion carves through space, recorded
directly by the head-mounted camera. Trajectory is a physically
grounded, fine-grained signal causally tied to outcome: the same motion
that produces the action also produces the trajectory, so the trajectory
carries enough of the action's structure to disambiguate it. 
CamFormer recently established this for the recognition setting:
trajectory already executed by the wearer, without any pixels, suffices
to recognize what someone has just done across a range of egocentric
activities \citep{xue2025seeing}. The signal that resolves ambiguity in
the \textbf{prediction} setting, however, is \textbf{future} trajectory:
the motion the wearer is about to execute, not the motion already
observed, and a signal that is not available at test time. A usable
model must therefore both predict plausible trajectories and commit to
action predictions consistent with them.

\vspace{-1.0em}
\paragraph{Two findings shape the design.}
The first is that future trajectory dominates language as a
conditioning signal for short-range egocentric prediction. Trained on
Ego-Exo4D under four conditioning regimes (none, language description
of the upcoming action, future trajectory, and both), a future-latent
predictor cuts validation $\ell_1$ roughly twenty times more under
trajectory than under language, and adding language on top of
trajectory adds nothing (Table~\ref{tab:motivation}). Shuffling the
trajectory input at test time hurts $\ell_1$ sharply; shuffling the
text input leaves it untouched: the signature of a content-sensitive
trajectory channel and a content-agnostic text channel.

\begin{table}[h]
\centering
\small
\setlength{\tabcolsep}{6pt}
\begin{tabular}{lcccccc}
\toprule
Conditioning & none & text & trajectory & text + trajectory & shuffled text$^*$ & shuffled trajectory$^*$ \\
\midrule
Best val $\ell_1$ ($\downarrow$) & 0.4773 & 0.4761 & 0.4572 & \textbf{0.4569} & 0.4575 & 0.4964 \\
Gain over \texttt{none}          & ---    & $-0.001$ & $-0.020$ & $-0.020$ & $-0.020$ & $+0.019$ \\
\bottomrule
\end{tabular}
\vspace{0.5mm}
\caption{Trajectory dominates language as a conditioning signal for
future-latent prediction on Ego-Exo4D ($H{=}1$, predicting one action
ahead). The first four columns are independent training regimes.
$^*$Shuffled-input columns are ablations on the best
\texttt{text+trajectory} checkpoint with the named input replaced by
input instance drawn from a different example (in-distribution but
unrelated to the target). Shuffling trajectory degrades val $\ell_1$
by $+0.039$; shuffling text degrades it by only $+0.0006$. The contrast
is the signature of a content-sensitive trajectory channel and a
content-agnostic text channel.}
\label{tab:motivation}
\end{table}

The second finding is that future trajectory is itself partially
predictable: start- and goal-state observations constrain plausible
motions, and a small bank of reference motions from training data
often covers the right answer. A model can therefore predict
candidates and use them as conditioning, recovering most of the gain
that ground-truth trajectory would have provided.

\vspace{-1.0em}
\paragraph{TrajPilot.}
TrajPilot instantiates these findings: it predicts a small set of
future-trajectory candidates from egocentric context and uses them to
pilot a causal predictor toward an action sequence. Predictions are
read out in an action-aligned embedding space (language shapes the
space but is never an input). Self-supervised video latents are strong
for perception but not organized around actions, and prove a poor
space to plan in (\S\ref{sec:method}). The alignment between
camera-trajectory latents and action embeddings lets TrajPilot combine
the geometry of the body's path with the semantics of the action it
indicates. Section~\ref{sec:exp} shows the trajectory channel is what
stabilizes long horizons. Across planning, anticipation, and event
prediction, TrajPilot is one model: planning is the standard
start-and-goal-conditioned setting, anticipation is the same model
with the goal masked, and event prediction is the same model applied
to a few frames of context.

\vspace{-1.0em}
\paragraph{Contributions.}
\begin{itemize}[leftmargin=1.5em, itemsep=2pt, topsep=2pt]
\item \textbf{Two empirical findings.} Future camera trajectory substantially outperforms language as a conditioning signal, and is itself partially predictable from the observed context, together making future trajectory a deployable conditioning signal at test time.
\item \textbf{TrajPilot, a trajectory-piloted predictor of human activity.} A causal model conditioned on (start, goal, predicted future trajectory) that reads predictions out in an action-aligned embedding space, with an inference-time scorer that decides when to trust a predicted trajectory and when to fall back. Section~\ref{sec:method} also diagnoses why a previously natural choice, predicting in self-supervised visual-latent space, fails for action-level planning.
\item \textbf{Procedural planning across four egocentric benchmarks.} On Ego-Exo4D atomic, Keystep, GoalStep, and EgoPER, TrajPilot outperforms strong VLM and structured-planner baselines (SCHEMA, PDPP, ViterbiPlanNet) re-trained under matched V-JEPA~2.1 features. The trajectory advantage widens with horizon, where the structured planners collapse, and holds when ground-truth trajectory is replaced with PI3 RGB-only pose estimation.
\item \textbf{One recipe, multiple tasks.} The same recipe applies
to goal-free anticipation on EPIC-Kitchens-100 (the action-aligned
readout token gives a small consistent gain over the visual-only
probe; trajectory headroom on this corpus is small), and predicts
whether a shot will score from a few frames of pre-shot context
(Ego-Exo4D basketball).
\end{itemize}

\section{Related Work}
\label{sec:related}

\vspace{-0.3em}
\paragraph{Ego world models for embodied agents.}
A line descending from Dreamer~\citep{hafner2023dreamer} predicts
future visual states conditioned on an agent's own action signal:
V-JEPA~2 / V-JEPA-2-AC~\citep{assran2025vjepa2} (joint-embedding
predictor post-trained on robot trajectory data, enabling
image-goal pick-and-place), NWM~\citep{bar2025nwm} (navigation
conditioned on locomotion controls), Cosmos~\citep{cosmos2024}
(pixels from text and image prefixes), GEM~\citep{gem2024}
(ego-trajectory, sparse features, human poses), PEVA~\citep{peva2025}
(whole-body pose), and EgoWM~\citep{egowm2026} (video diffusion
fine-tuned into action-conditioned simulators for 3-DoF mobile
robots and 25-DoF humanoid joints). All condition on the agent's
actuator state and predict what the agent will see. We instead
observe a person passively, condition on the wearer's trajectory,
and predict their activity, not pixels.

\vspace{-1.0em}
\paragraph{Procedural planning in instructional video.}
Procedural planning~\citep{chang2020procedure}, benchmarked on
COIN~\citep{tang2019coin} and CrossTask~\citep{zhukov2019crosstask},
predicts a sequence of action steps from start and goal observations.
Approaches include diffusion planners
(PDPP~\citep{pdpp2023} with classifier-free
guidance~\citep{ho2022cfg}), state-change-aware
planners (SCHEMA~\citep{schema2024}), and frozen-encoder diffusion
transformers (VEDiT~\citep{vedit2025}). JEPA-style entrants
descended from I-JEPA~\citep{assran2023ijepa} and
V-JEPA~\citep{bardes2024vjepa} select actions whose predicted latent
minimizes goal distance: V-JEPA-2-AC in V-JEPA latent space, and
GeoWorld~\citep{geoworld2026} in hyperbolic latent space with CEM
over geodesics. None use camera trajectory. We add trajectory as a
planning input and diagnose (\S\ref{sec:method}) why CEM in
self-supervised visual-latent space fails for action-level planning.

\vspace{-1.0em}
\paragraph{Trajectory as a semantic signal.}
CamFormer~\citep{xue2025seeing} shows that camera trajectories
alone, without pixels, suffice to recognize what someone is doing
across egocentric activities. We extend this from recognition to
prediction: we adapt their trajectory encoder as our alignment
encoder (\S\ref{sec:method}) and use future trajectory as a
planner's control signal. EgoDistill~\citep{egodistill2023} uses
head-motion-from-IMU as a distillation target for efficient video
understanding; we instead treat trajectory as a first-class
conditioning input the model predicts and consumes at inference.

\vspace{-1.0em}
\paragraph{Text-embedding-target prediction.}
A recent line predicts language-space targets in place of tokens or
pixels: VL-JEPA~\citep{chen2025vljepa} predicts continuous text
embeddings as a general-purpose vision-language JEPA, and
VLWM~\citep{vlwm2025} predicts trajectories of language-described
actions and world-state changes with a self-supervised critic for
reflective system-2 planning. We share the language-aligned readout
but our model never \emph{reads} language: it predicts in
EgoVLPv2~\citep{lin2022egovlp} space without language input, and
prediction is structured around a (start, mid$_1\dots$mid$_h$, goal)
skeleton rather than free-form text, with trajectory piloting the
per-step rollout.

\vspace{-1.0em}
\paragraph{Action anticipation in egocentric video.}
EPIC-Kitchens-100~\citep{damen2022rescaling} anticipation has driven
dedicated models from AVT~\citep{girdhar2021anticipative} to recent
joint-embedding predictors V-JEPA~2~\citep{assran2025vjepa2} and
V-JEPA~2.1~\citep{vjepa21}. We obtain anticipation from our planner by masking the goal token at
inference, rather than training a separate model.

\section{Method}
\label{sec:method}

\subsection{Problem setup}
\label{sec:method:setup}

We consider near-future prediction from egocentric video: given a
short clip of egocentric context, predict an aspect of the immediate
future. Three concrete instantiations recur in this paper. 
In \emph{procedural planning}, context is a start clip and a goal clip,
and the prediction is the middle of an $H$-step plan
$[\text{Start}, \text{Mid}_1, \ldots, \text{Mid}_h, \text{Goal}]$ with
$h = H - 2$ middle actions to fill in, drawn from an open-vocabulary
action bank ($H \in \{3,\ldots,8\}$, so $h \in \{1,\ldots,6\}$).
In \emph{action anticipation}, context
is the start clip alone with no goal, and the prediction is the
upcoming action sequence. In \emph{event prediction}, context is a
few frames and the prediction is a single discrete outcome.

All three instantiations share the same internal decomposition. A
\emph{predictor} consumes the available context and produces an
intermediate representation $z_1, \ldots, z_h$ at each future step.
A \emph{readout} maps each $z_t$ to a discrete action or outcome
label. The central design questions are: in what space should the
predictor operate, and what additional signal should condition it?

\subsection{Background: latent-space planning}
\label{sec:method:background}

The natural baseline for this kind of prediction is to operate
entirely in the latent space of a strong self-supervised video
encoder (e.g., V-JEPA~2.1~\citep{vjepa21}). A frozen encoder maps
the start clip and the goal clip to start and goal latents
$z_\text{start}$, $z_\text{goal}$. A learned predictor takes
$(z_\text{start}, z_\text{goal})$ and produces predicted intermediate
latents $\hat{z}_1, \ldots, \hat{z}_h$. A candidate plan is scored
by how close its predicted final latent $\hat{z}_h$ is to
$z_\text{goal}$ under an $\ell_1$ objective, and the action label
at each step is read out from $\hat{z}_t$ via a separately trained
classification head. To select among candidates, this recipe
typically searches over action sequences with the cross-entropy
method (CEM), which iteratively refines a proposal distribution
toward sequences whose predicted final latent lands close to
$z_\text{goal}$. GeoWorld~\citep{geoworld2026} is a recent
instance of this recipe in a hyperbolic latent space, with CEM
search over geodesics.

In this paper, we test a stronger version of the recipe by giving
the predictor an additional conditioning signal: future camera
trajectory. We compare three settings, all sharing the same
predictor architecture and the same $\ell_1$-to-goal scoring
objective, with each setting trained for its intended test-time
use. \textbf{No-Traj} is trained without trajectory tokens and runs
without any trajectory input at inference. \textbf{Oracle} is
trained with ground-truth trajectory tokens and fed the ground-truth
future trajectory at inference, providing an upper bound on what
trajectory conditioning can buy in this latent space. \textbf{CEM}
uses the same trained predictor as Oracle, but at inference samples
candidate future trajectories, rolls each through the predictor, and
selects the one whose predicted final latent lands closest to
$z_\text{goal}$ under $\ell_1$, refining the proposal distribution
with the cross-entropy method.

\subsection{Diagnostic: latent-space planning fails}
\label{sec:method:diagnostic}

\begin{wrapfigure}{r}{0.45\linewidth}
\centering
\includegraphics[width=\linewidth]{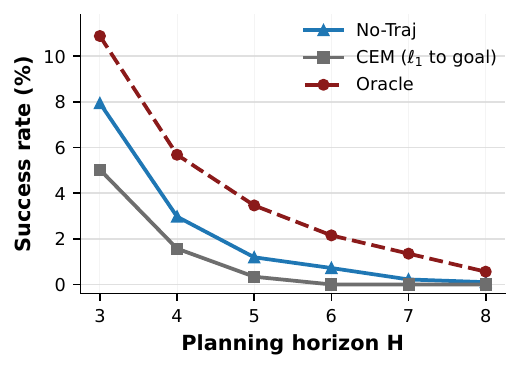}
\caption{Latent-space planning fails on Ego-Exo4D atomic-action
planning. CEM (searching trajectory candidates by $\ell_1$ to the
goal latent) underperforms No-Traj at every horizon; Oracle
(ground-truth trajectory) shows trajectory carries substantial
signal. Full mean accuracy in Figure~\ref{fig:cem_full_mean}.}
\label{fig:cem_vs_none}
\end{wrapfigure}

We instantiated this recipe with V-JEPA~2.1 as the encoder and a
10-layer transformer as the predictor, and evaluated all three
settings on Ego-Exo4D atomic-action planning across horizons $H = 3$
to $H = 8$. Figure~\ref{fig:cem_vs_none} shows the result.

Two patterns emerge. First, Oracle substantially outperforms
No-Traj at every horizon: trajectory carries useful signal in this
latent space, and a predictor with access to it makes meaningfully
better predictions than one without. Second, and damagingly, CEM
underperforms even No-Traj at every horizon. Searching for the
best trajectory under the $\ell_1$-to-goal objective picks
\emph{worse} trajectories than ignoring trajectory entirely. The
search is not just imperfect; it is actively harmful.

The cause is straightforward. Self-supervised video encoders are
trained to predict future frames, not to separate actions: their
latent space is organized by visual continuity, not by what someone
is doing. A direct probe of the structure confirms this: two clips from the
same recording are closer in V-JEPA space than two clips of the
same action from different recordings
(Table~\ref{tab:vjepa_geometry}), and ground-truth trajectories
rarely move monotonically closer to the goal in latent $\ell_1$ at
long horizons (Table~\ref{tab:monotonicity}). $\ell_1$ distance
in this space therefore distinguishes obviously-different futures
but cannot tell apart candidate trajectories that lead to visually
similar futures, which is exactly what CEM needs to do. Oracle
sidesteps this issue by skipping CEM altogether: given the
ground-truth trajectory directly as input, the predictor never needs
$\ell_1$-to-goal scoring to find it. The harm is specific to the
search step.

\begin{figure*}[t]
\centering
\includegraphics[width=\linewidth]{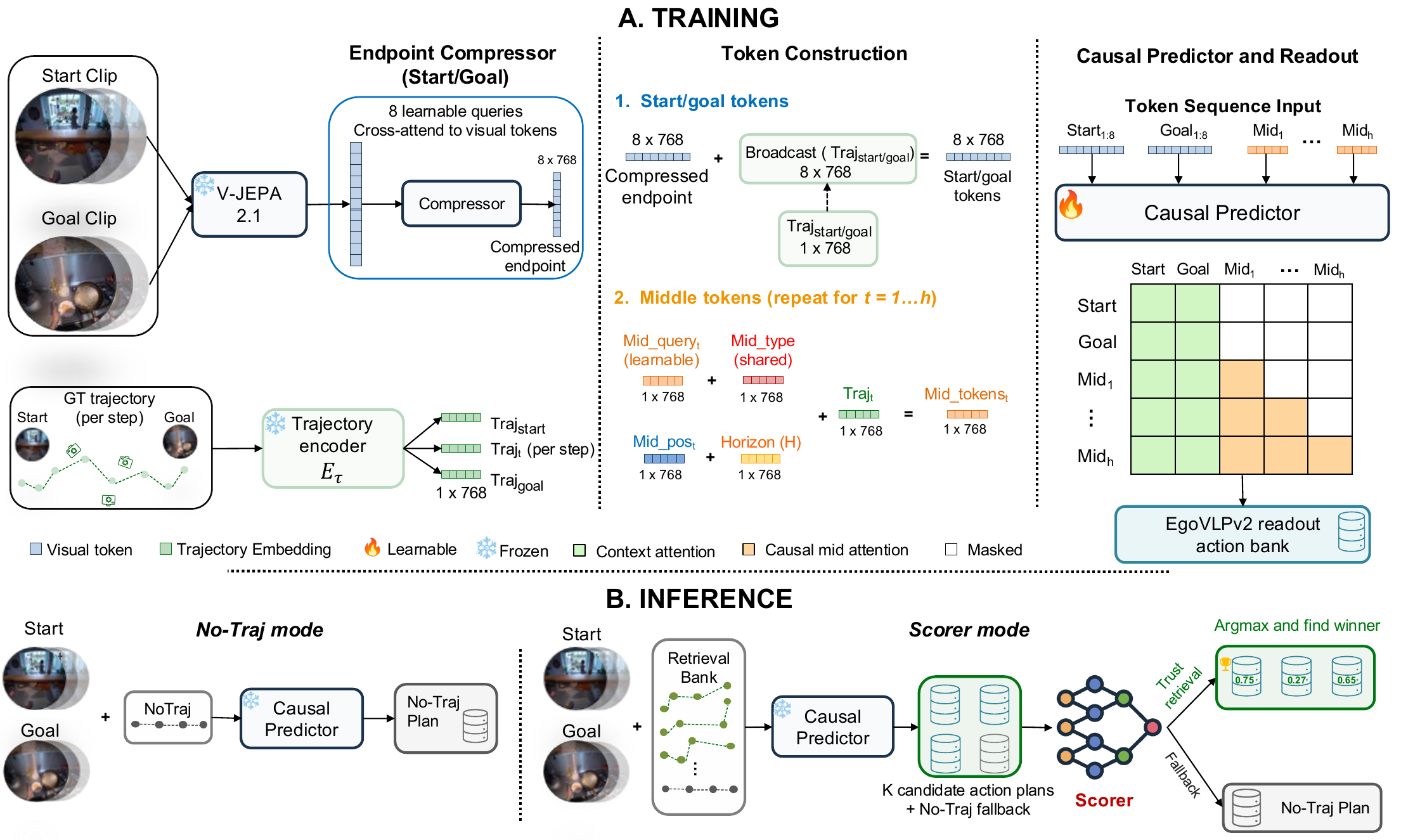}
\caption{TrajPilot architecture overview.
\textbf{(A) Training:} V-JEPA context features and trajectory
embeddings from $E_\tau$ feed an additive token construction read by
a causal predictor that scores against the EgoVLPv2 action bank
(\S\ref{sec:method:stage1}--\ref{sec:method:stage2}).
\textbf{(B) Inference:} \emph{No-Traj mode} runs the predictor with
a zero trajectory input; \emph{Scorer mode} retrieves $K$ candidate
trajectories, runs the predictor over each in parallel, and selects
via the gate-then-rank scorer, with a no-trajectory fallback
(\S\ref{sec:method:stage3}).}
\label{fig:arch}
\vspace{-0.5em}
\end{figure*}

Three design choices follow, realized in the next three
subsections: predict in an action-aligned space rather than
V-JEPA latent space (\S\ref{sec:method:stage1}), encode
trajectory into that same space (\S\ref{sec:method:stage2}),
and predict trajectory at inference (\S\ref{sec:method:stage3}).
Figure~\ref{fig:arch} summarizes the resulting architecture.

\subsection{Trajectory--action alignment}
\label{sec:method:stage1}

Given the readout space chosen, we want trajectory to live in the
same space, so that the predictor can attend to a single
action-aligned representation. Following CamFormer
\citep{xue2025seeing}, we encode each segment's 16-knot relative
6-DoF trajectory $u \in \mathbb{R}^{16 \times 6}$ with a small
transformer (4 layers, width 128, mean-pooled), trained from scratch
under a contrastive objective that aligns its outputs
with EgoVLPv2 text embeddings rather than with action class labels.
Concretely, for a batch of $B$ clips, let $z_i = E_\tau(u_i) \in
\mathbb{R}^{d}$ be the trajectory embedding and $e_i \in \mathbb{R}^{d}$
the frozen EgoVLPv2 embedding of the clip's action description, both
$\ell_2$-normalized; let $t_i$ be the action's text ID. We compute
similarity logits $\ell_{ij} = \tau \, z_i^\top e_j$ with learnable
temperature $\tau$, and define a duplicate-aware positive mask
$M_{ij} = \mathbf{1}[t_i = t_j]$. The loss is the symmetrized
multi-positive cross-entropy
\begin{align*}
\mathcal{L}_{\text{align}} &\;=\;
\tfrac{1}{2}\bigl(\mathrm{mpCE}(\ell, M) + \mathrm{mpCE}(\ell^\top, M^\top)\bigr), \\
\mathrm{mpCE}(\ell, M) &\;=\; -\tfrac{1}{B}\sum_{i}
\Bigl[
\log\!\sum_{j: M_{ij}=1} e^{\ell_{ij}}
\;-\;
\log\!\sum_{j} e^{\ell_{ij}}
\Bigr],
\end{align*}
treating all paraphrases of the same action as positives rather than
as negatives of one another. Without this correction, standard
one-hot InfoNCE penalizes the model for putting paraphrases close
together.

After this alignment, $E_\tau$ produces per-step trajectory embeddings that
live in EgoVLPv2 space and cluster by action semantics. The alignment encoder is frozen for the
remainder of the pipeline.

\subsection{Causal predictor in action-aligned space}
\label{sec:method:stage2}

The predictor is a causal transformer \citep{vaswani2017attention}
with learned additive embeddings for token type, step position within the plan, and plan horizon,
that takes the input sequence
$[\, \text{Start}_{1:8}, \text{Goal}_{1:8}, \text{Mid}_1, \ldots, \text{Mid}_h \,]$
of length $16 + h$ and predicts the EgoVLPv2 embedding of each mid-step in a single
forward pass under a causal mask. Start and Goal are each
represented as $8$ compressed tokens produced by a bank of $8$
learnable query tokens that cross-attend into the V-JEPA~2.1 token
sequence of the corresponding clip; the per-segment trajectory
embedding $z_{\text{start}}, z_{\text{goal}}$ from $E_\tau$ is
projected and added into these compressed context tokens.
Each mid-token sums five learnable inputs: the trajectory embedding
from $E_\tau$ for that step, a query token (the slot the model
writes its prediction into), a step-position embedding, a
horizon embedding, and a mid-type embedding that distinguishes
mid tokens from start/goal context tokens.
A structured attention mask makes Start/Goal
bidirectional, mids causal, and one-way from context into mids
(context is blind to mids), so Mid$_j$ cannot leak into Mid$_i$ for
$i < j$.

The training loss combines a cosine-alignment term that pulls each
predicted embedding toward its ground-truth EgoVLPv2 target, and a
symmetric multi-positive InfoNCE term (\S\ref{sec:method:stage1}) over the pooled set of all
predicted tokens (Start, Mid$_1, \ldots,$ Mid$_h$, Goal) in the
batch:
\[
\mathcal{L}_{\text{pred}} \;=\;
\sum_{r \in \{\text{S}, \text{M}, \text{G}\}}
\bigl(1 - \cos(\hat{e}^r,\, e^r)\bigr)
\;+\;
\lambda \, \mathcal{L}_{\text{InfoNCE}}^{\text{pool}},
\]
with $\lambda = 0.5$ and the InfoNCE term using the same
duplicate-aware positive mask as the alignment loss, applied across
a single pool that mixes Start, Mid, and Goal predictions to prevent
role-specific shortcuts. Trajectory dropout ($p = 0.1$)
regularizes against over-reliance on the trajectory channel.

\subsection{Trajectory retrieval and gate-then-rank scorer}
\label{sec:method:stage3}

At inference, the test-time trajectory is unobserved. The
\emph{gate-then-rank scorer} fills this gap. For each test input
$(x_s, x_g)$, we retrieve the top-$K{=}64$ training-set trajectories
by Start/Goal endpoint cosine and roll each through the frozen
trajectory-conditioned predictor; in parallel, the frozen No-Traj
predictor produces a fallback prediction. A scorer transformer reads
the per-candidate predictions, the fallback prediction, retrieved
trajectory features, and a query vector pooled from Start/Goal, and
emits two outputs: a binary \emph{gate} logit (does any retrieved
candidate beat the fallback?) and per-candidate \emph{rank} scores
used when the gate is open. At test time, if the gate fires negative,
return the No-Traj prediction; else return the highest-rank retrieved
candidate (Figure~\ref{fig:scorer} in
Appendix~\ref{app:scorer_mechanism}).
The top-$64$ retrieved pool is rich: an oracle picking the
best candidate from the pool roughly doubles per-step accuracy
relative to No-Traj at $H{=}5$
(Table~\ref{tab:retrieval_headroom}),
so the bottleneck is selection, not retrieval. The scorer is trained
with a retrieval-aligned utility (precise form in
Appendix~\ref{app:scorer_loss}). At $H{=}3$ it falls back to No-Traj
for roughly $81\%$ of test inputs (where Start/Goal alone determine
the plan); at $H{=}8$, only $\sim 18\%$
(Appendix~\ref{app:selector_behavior}).

\section{Experiments}
\label{sec:exp}

We evaluate \ours{} on procedural planning and goal-free anticipation.
Throughout, the visual encoder (frozen \vjepag{}) and alignment encoder
$E_\tau$ are held fixed; the causal predictor and scorer are trained
once on open-vocabulary atomic data and either evaluated as-is or
fine-tuned per benchmark with a shared recipe.

\vspace{-0.5em}
\subsection{Setup}
\label{sec:exp:setup}

\vspace{-0.3em}
\paragraph{Datasets.}
We evaluate on four egocentric benchmarks: \textbf{\egoexo{} atomic}
\citep{egoexo4d}, an open-vocabulary atomic-action benchmark whose
$8{,}472$ labels are short action descriptions (e.g. ``cut the carrot
with a knife''); \textbf{\egoexo{} Keystep} \citep{egoexo4d}, a
closed-vocabulary keystep benchmark with $375$ labels and per-clip
scenario tags; \textbf{Ego4D GoalStep} \citep{song2024goalstep}, a
long-take closed-vocabulary procedural-planning benchmark; and
\textbf{EgoPER} \citep{leeprocedural2024}, an egocentric
procedural-error benchmark with $57$ labels. Dataset statistics and
splits are in Appendix~\ref{app:implementation:datasets}.
\vspace{-1.0em}
\paragraph{Variants of \ours{} compared.}
\textbf{No-Traj} is a separate checkpoint trained from scratch with no
trajectory input. \textbf{Scorer} and \textbf{Oracle} share a single
trajectory-trained checkpoint and differ only at inference. Scorer is
the gate-then-rank pipeline of \S\ref{sec:method:stage3}: it retrieves
$K$ candidate trajectories from training data and either picks one or
falls back to the no-trajectory branch ($K$ values per benchmark in Appendix~\ref{app:implementation:closed-vocab}). This is the actual test-time
\ours{}, since future trajectory is not observed. \textbf{Oracle}
feeds the predictor the ground-truth middle trajectory and is reported
only as an upper bound.\vspace{-1.0em}
\paragraph{Metrics.}
For open-vocabulary atomic, middle-step retrieval R@1/R@5 (M@1, M@5),
exact middle-sequence match (MSeq), and full-sequence counterparts
(F@1, F@5, FSeq). For closed-vocabulary benchmarks, mid and full mean
accuracy (Mid/Full $\macc$), exact-sequence success rate ($\sr$),
full-sequence mIoU, and Levenshtein edit distance (lower is better).
\vspace{-1.0em}
\paragraph{Baselines.}
For open-vocabulary atomic planning and anticipation, we compare
against two VLM baselines. \textbf{Qwen-ZS} prompts
Qwen3-VL-32B zero-shot to generate the missing action sequence from
the visual context; the generated string is matched into the
$8{,}472$-label action bank by \egovlp{} cosine similarity.
\textbf{Qwen-SFT$+$LLM} is a two-stage version: a
supervised-finetuned Qwen3-VL is run on the start and goal clips alone
to predict the start and goal action labels (reaching $43.6\%$ R@1 on
this endpoint-recognition subtask, see
Table~\ref{tab:qwen3vl_endpoint_cache}); Qwen3-30B then receives
those predicted endpoint labels plus the action-name bank and fills
the middle sequence. The two-stage version isolates how much of the
\ours{} gap is due to weak VLM recognition versus the harder
middle-step inference (full prompts and SFT recipe in
Appendix~\ref{app:vlm_baseline_details}). For closed-vocabulary planning, we
compare against SCHEMA~\citep{schema2024}, PDPP~\citep{pdpp2023}, and
\viterbinet{}~\citep{viterbiplannet}, 
retrained on matched \vjepag{} features and run with its native
decoder.

\vspace{-0.5em}
\subsection{Atomic-action planning on \egoexo{}}
\label{sec:exp:atomic}

\begin{figure*}[t]
\centering
\includegraphics[width = \linewidth]{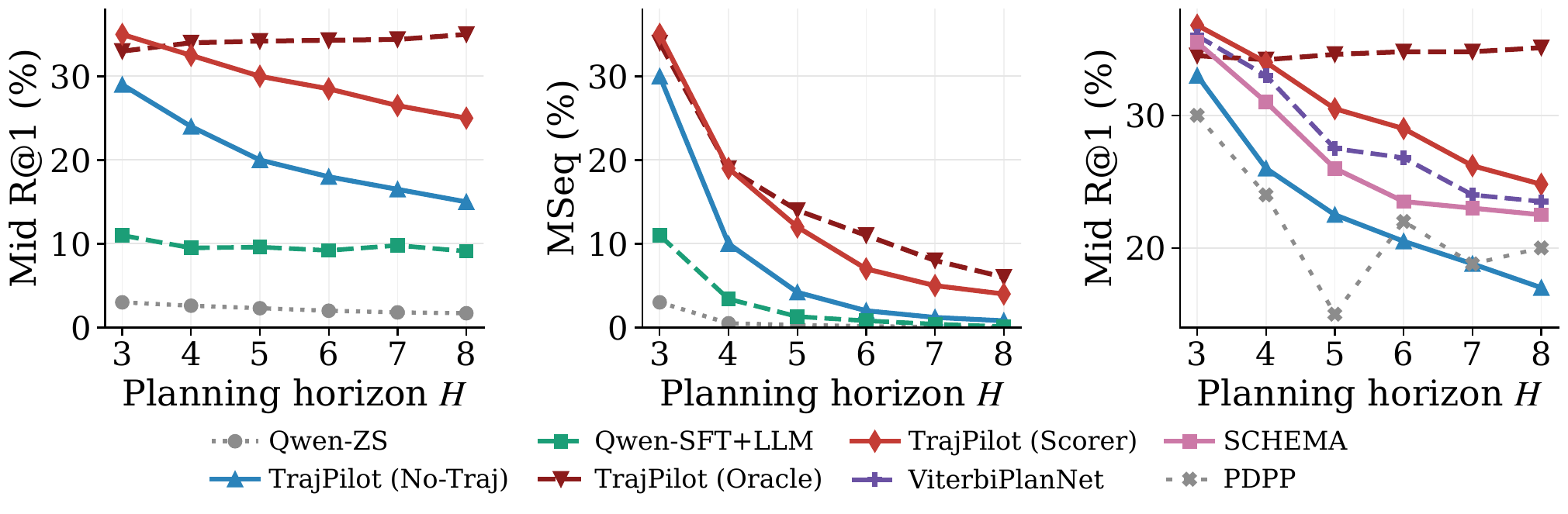}
\caption{Atomic-action planning on \egoexo{}. \emph{Left and middle:}
open vocabulary ($|\mathcal{V}|{=}8{,}472$); mid-step retrieval (M@1)
and exact mid-sequence match (MSeq) versus planning horizon $H$;
baselines are VLMs. \emph{Right:} closed vocabulary
($|\mathcal{V}|{=}1{,}165$); Mid R@1; baselines are structured
procedural planners. Oracle (dashed) is a diagnostic upper bound that
uses ground-truth middle trajectory at inference. Per-horizon
open-vocabulary detail in Table~\ref{tab:atomic_results_full}.}
\label{fig:atomic_planning}
\end{figure*}

We evaluate procedural planning on \egoexo{} atomic under two
vocabularies: the open $8{,}472$-label bank, with all methods scored
against the same \egovlp{} text bank; and a closed $1{,}165$-label
collapse that classical procedural planners can be trained on. \ours{}
is the same model in both regimes, with only the readout bank
restricted at evaluation; SCHEMA, PDPP, and \viterbinet{} are
retrained on matched \vjepag{} features and run with their native
decoders, and apply only in the closed-vocabulary regime. The
closed-vocabulary protocol is structurally favourable to the
baselines: their capacity is matched to the $1{,}165$-class graph,
while \ours{} is trained for the larger open bank. Results across
horizons $H \in \{3,\ldots,8\}$ are in Figure~\ref{fig:atomic_planning}.

\vspace{-1.0em}
\paragraph{Open vocabulary.}
Three findings.
\emph{1) Both VLMs fail at the open-vocabulary task.}
Qwen-ZS retrieves poorly ($1$--$3\%$ M@1, far below the trajectory
variants), and even Qwen-SFT$+$LLM reaches only $9$--$11\%$ M@1, leaving
a $+22$--$24$ pp gap below \ours{} (Scorer).
\emph{2) No-Traj collapses with horizon} ($29.4 \to 15.1\%$ M@1 from
$H{=}3$ to $H{=}8$): without trajectory conditioning, Start and Goal
alone leave the plan under-determined as the horizon stretches.
\emph{3) The Scorer recovers most of the No-Traj-to-Oracle gap}
without any test-time trajectory supervision, and the recovery widens
with horizon. At $H{=}8$, the Scorer reaches $24.7\%$ M@1 against an
Oracle upper bound of $34.4\%$, while No-Traj sits at $15.1\%$.

\vspace{-1.0em}
\paragraph{Closed vocabulary.}
\ours{} (Scorer) wins on the $1{,}165$-class graph too:
sample-weighted overall, Mid R@1 $\mathbf{30.6}$ / Mid Seq
$\mathbf{15.2}$, ahead of \viterbinet{} ($26.5 / 15.1$), SCHEMA
($25.6 / 14.5$), and PDPP ($20.4 / 10.3$); Oracle reaches
$34.8 / 16.2$. The gap widens with horizon
(Figure~\ref{fig:atomic_planning}, right). Closed-vocab decoders win
M@5, where the transition prior helps most, but \ours{} leads on
Mid R@1 and Mid Seq while also supporting the open bank.

\vspace{-0.5em}
\subsection{Closed-vocabulary planning across procedural benchmarks}
\label{sec:exp:coarse}
\vspace{-0.3em}

We transfer the same backbone to \egoexo{} Keystep,
Ego4D GoalStep, and EgoPER, fine-tuning the predictor with a
per-benchmark classifier head and leaving goal dropout, attention,
and the scorer unchanged (per-benchmark hyperparameters in
Table~\ref{tab:closed_vocab_hparams}, per-horizon detail and
edit-distance results in Table~\ref{tab:coarse_planning_full}).
\ours{} (\retrtraj) wins $7$ of $9$ (dataset, metric) cells against
the structured-planner baselines (Table~\ref{tab:coarse_planning_overall};
\viterbinet{} edges us on Keystep SR and GoalStep mAcc, where the
transition prior helps most). The atomic-pretrained
backbone transfers without benchmark-specific encoder
retraining: \ours{} (\notraj) alone matches or exceeds the strongest
prior planner on Keystep mAcc/mIoU, GoalStep mIoU, and all three
EgoPER metrics. The \retrtraj{}-vs-\notraj{} gap is under $1$ pp on
most cells, versus $5$--$10$ pp in open-vocab, reflecting that the
smaller label space already constrains the prediction; the
trajectory advantage is largest where the vocabulary is least
restrictive.

\begin{table*}[t]
\centering
\small
\setlength{\tabcolsep}{4pt}
\caption{Closed-vocabulary procedural planning, average 
across horizons $H{=}3,\ldots,8$. Test sets: \egoexo{} Keystep
($|\mathcal{V}|{=}375$), Ego4D GoalStep ($|\mathcal{V}|{=}310$), and EgoPER ($|\mathcal{V}|{=}57$). SR is full-sequence exact match; mAcc is
mid-step mean accuracy; mIoU is full-sequence IoU. All metrics in
percent (higher is better). We report mid mAcc rather than full mAcc because the start and goal
endpoints are observed, so any effect of trajectory conditioning is
concentrated in the middle steps. Best non-oracle method per (dataset,
metric) is bolded. \ours{} (\oracletraj) is a diagnostic upper bound
that uses ground-truth middle trajectory at inference. Per-horizon
detail and edit-distance results in
Table~\ref{tab:coarse_planning_full}.}
\label{tab:coarse_planning_overall}
\begin{tabular}{lccc ccc ccc}
\toprule
& \multicolumn{3}{c}{Keystep} & \multicolumn{3}{c}{GoalStep} & \multicolumn{3}{c}{EgoPER} \\
\cmidrule(lr){2-4} \cmidrule(lr){5-7} \cmidrule(lr){8-10}
Method & SR & mAcc & mIoU & SR & mAcc & mIoU & SR & mAcc & mIoU \\
\midrule
SCHEMA                & $2.19$ & $13.39$ & $23.86$ & $0.48$ & $5.17$ & $5.49$ & $23.88$ & $45.27$ & $56.00$ \\
PDPP                  & $1.75$ & $14.56$ & $23.94$ & $0.21$ & $4.90$ & $8.57$ & $70.40$ & $81.49$ & $88.97$ \\
\viterbinet{}         & $\mathbf{2.98}$ & $15.26$ & $26.78$ & $0.63$ & $\mathbf{7.25}$ & $9.77$ & $53.10$ & $73.43$ & $80.97$ \\
\ours{} (\notraj)     & $2.56$ & $16.63$ & $29.55$ & $0.66$ & $6.79$ & $10.12$ & $70.17$ & $82.56$ & $90.68$ \\
\ours{} (\retrtraj)   & $2.64$ & $\mathbf{16.75}$ & $\mathbf{29.58}$ & $\mathbf{0.82}$ & $6.86$ & $\mathbf{10.63}$ & $\mathbf{70.59}$ & $\mathbf{82.78}$ & $\mathbf{90.76}$ \\
\midrule
\ours{} (\oracletraj) & $2.98$ & $20.11$ & $29.52$ & $0.77$ & $7.77$ & $11.14$ & $72.35$ & $84.04$ & $89.97$ \\
\bottomrule
\end{tabular}
\end{table*}

\vspace{-0.7em}
\subsection{Robustness to RGB-only camera-pose estimation}
\label{sec:exp:source-shift}
\vspace{-0.3em}
Aria 6-DoF pose is not always available: most egocentric video has
only RGB. We test \ours{} with PI3~\citep{pi3} pose estimated from
RGB on Keystep (Table~\ref{tab:pi3_robustness}), comparing matched
training/eval sources (\textbf{Aria$\to$Aria}, \textbf{PI3$\to$PI3}),
the source-mismatch failure mode (\textbf{Aria$\to$PI3}), and a mixed
checkpoint (\textbf{Aria$+$PI3$\to$PI3}) trained on the union and
evaluated on PI3. All settings share the same frozen $E_\tau$; only
the predictor's trajectory input distribution differs.
PI3$\to$PI3 matches Aria$\to$Aria within $\pm 0.2$ Mid $\macc$
($19.65$ vs.\ $19.47$): the frozen $E_\tau$ transfers to RGB-estimated
trajectories without retraining, so adaptation is on the predictor side.
The Aria$\to$PI3 row shows this adaptation is necessary: an
Aria-only predictor drops to $12.65$ Mid $\macc$ on PI3 ($-6.82$ pp).
A mixed Aria$+$PI3 checkpoint 
matches both single-source variants within $\pm 0.2$ Mid/Full $\macc$.

\vspace{-0.5em}
\subsection{Goal masking unlocks anticipation from the same checkpoint}
\label{sec:exp:anticipation}
\vspace{-0.3em}
Action anticipation differs from procedural planning only in that the
goal endpoint is unobserved. To test whether the same model serves
both tasks, we mask the goal token at inference and compare two
checkpoints: a \emph{no-dropout} checkpoint trained as a standard
goal-conditioned planner, and a \emph{goal-dropout} checkpoint trained
with per-sample goal dropout at $p_{\text{goal}}{=}0.5$
(\S\ref{sec:method:stage2}), which sees a masked goal $50\%$ of the
time during training. Goal dropout is at worst neutral on standard
full-goal planning (overall Future R@1: $33.8 \to 34.6$ with
trajectory, $20.8 \to 23.2$ without; full per-horizon table in
Table~\ref{tab:goal_dropout_fullgoal}). With the goal masked at
inference, dropout lifts overall Future R@1 from $30.7$ to $33.9$
with trajectory ($+3.2$ pp) and from $11.2$ to $18.7$ without
($+7.5$ pp), shrinking the planning-to-anticipation drop with
trajectory from $-3.1$ to $-0.7$ pp. The result is a single
checkpoint that serves planning and goal-free anticipation without
task-specific finetuning (per-horizon detail in
Table~\ref{tab:goal_dropout_anticipation_full}).

\begin{figure}[!t]
\centering
\begin{minipage}[t]{0.63\linewidth}
\vspace{0pt}
\centering
\includegraphics[width=\linewidth]{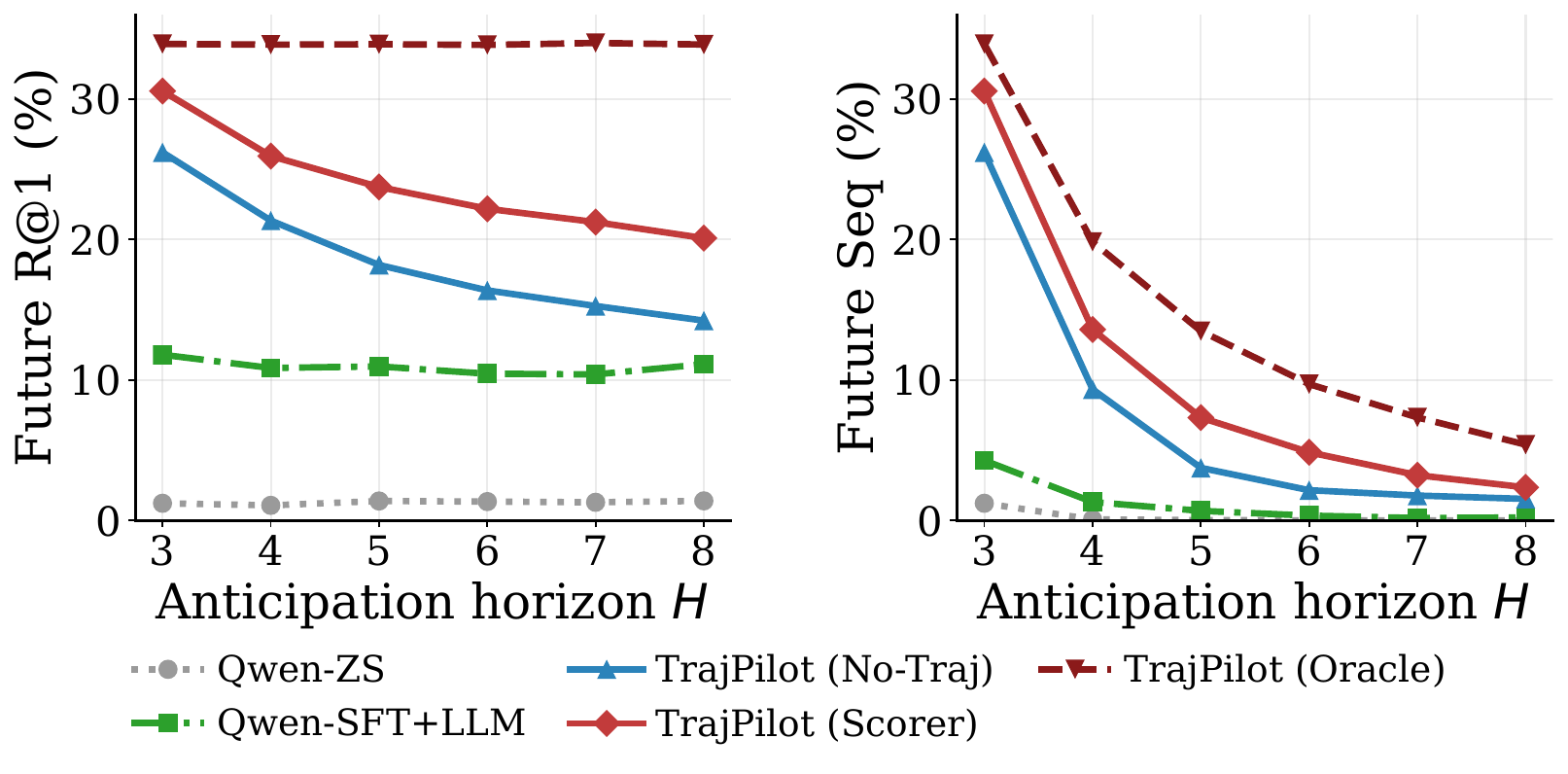}
\caption{Open-vocabulary atomic anticipation on \egoexo{} atomic
(test split, $8{,}472$-label atomic action bank), goal
removed at inference. Future-step retrieval (Future R@1, left) and
exact future-sequence match (Future Seq, right) versus anticipation
horizon $H$. Oracle (dashed) is a diagnostic upper bound that uses
ground-truth middle trajectory at inference. Per-horizon detail across
all six metrics in Table~\ref{tab:atomic_anticipation_full}.}
\label{fig:atomic_anticipation}
\end{minipage}\hfill
\begin{minipage}[t]{0.34\linewidth}
\captionof{table}{Trajectory-source robustness for \ours{}
(\oracletraj) on Keystep, sample-weighted across $H{=}3$--$8$
(train$\to$eval; higher is better). Aria$\to$PI3 is a source-mismatch
failure mode; Aria$+$PI3$\to$PI3 trains on the union and evaluates on PI3.}
\label{tab:pi3_robustness}
\vspace{0pt}
\centering
\vspace{0.5em}
\footnotesize
\setlength{\tabcolsep}{2pt}
\begin{tabular}{lcc}
\toprule
Setting & Mid $\macc$ & Full $\macc$ \\
\midrule
Aria$\to$Aria       & $19.47$ & $24.29$ \\
PI3$\to$PI3         & $19.65$ & $24.23$ \\
Aria$\to$PI3        & $12.65$ & $18.96$ \\
Aria$+$PI3$\to$PI3  & $19.53$ & $24.37$ \\
\bottomrule
\end{tabular}
\end{minipage}
\end{figure}

\paragraph{Atomic anticipation against VLM baselines.}
Using the goal-dropout checkpoint, we compare atomic anticipation
against the same VLM baselines as \S\ref{sec:exp:atomic}. The scorer
indexes by start only (top-$64$); the rest of the pipeline is
unchanged. Results are in Figure~\ref{fig:atomic_anticipation}.
Two observations.
\emph{1) \ours{} (Scorer) is the strongest test-time method on
Future R@1 at every horizon}, beating the best VLM by $+13$ to $+19$
pp.
\emph{2) Oracle stays nearly flat at $\sim 33.9$ Future R@1} across
$H{=}3,\ldots,8$, identical to its planning value with the goal
observed (cf.\ Table~\ref{tab:atomic_results_full}). With correct
trajectory the goal becomes nearly redundant: trajectory carries
enough of the procedural intent to disambiguate the future on its
own.

\paragraph{Cross-corpus transfer.}
The recipe applies to EPIC-Kitchens-100
anticipation~\citep{damen2022rescaling} on the V-JEPA~2
attentive-probe protocol~\citep{vjepa21}: injecting the predictor's
action-aligned readout token into the probe lifts every metric over
the visual-only baseline ($+0.5$ Action R@5, $+1.35$ Action Acc);
adding retrieved trajectory does not further improve Action R@5, and
even ground-truth trajectory adds only $+0.78$ pp Action R@5, so
trajectory headroom on this corpus is small. Detail in
Appendix~\ref{app:ek100}.

\vspace{-0.4cm}
\section{Conclusion}
\vspace{-0.2cm}
\label{sec:conclusion}

Two findings shape this work. Future camera trajectory carries enough
of how the body moves to determine outcome, far more than language
does, and is itself predictable from observed context. TrajPilot
predicts candidate trajectories from start and goal context and pilots
a causal predictor in an action-aligned embedding space. Across four
egocentric planning benchmarks it outperforms VLM and structured-planner
baselines under matched conditions, with the trajectory advantage
widening at long horizons and holding under RGB-only pose estimation;
the same checkpoint serves goal-free anticipation. Two limits remain:
the deployable Scorer trails the Oracle at long horizons (with a rich
retrieved pool, the bottleneck is selection, not coverage), and
trajectory adds little on corpora where long-tail object recognition,
rather than motor disambiguation, drives the task (EK100). How you
move tells what you'll do, and predicting that motion is enough to
predict the rest.

\bibliographystyle{plainnat}
\bibliography{references}

\appendix

\newpage
\section{Implementation details}
\label{app:implementation}

\subsection{Datasets}
\label{app:implementation:datasets}

\vspace{-1.0em}
\paragraph{\egoexo{} atomic.}
We use the open-vocabulary \egoexo{} atomic-action benchmark
\citep{egoexo4d}: $151.8$K short atomic-action segments across
$2{,}549$ takes and $71$ task scenarios. The split is take-disjoint,
with $140{,}413$ train segments from $2{,}237$ train takes and
$11{,}425$ validation segments from $312$ validation takes. Planning
manifests for horizons $H \in \{3,\ldots,8\}$ are constructed as
consecutive atomic-action windows within each split. Predictions are
scored against the EgoVLPv2 atomic text bank with
$|\mathcal{V}_{\text{atomic}}| = 8{,}472$ open-vocabulary labels.
\vspace{-1.0em}
\paragraph{\egoexo{} Keystep.}
We use the \egoexo{} Keystep actions \citep{egoexo4d}: $18{,}091$
segment clips, split take-disjointly into $13{,}819$ train clips and
$4{,}272$ validation clips. The label space contains
$|\mathcal{V}_{\text{keystep}}| = 375$ closed labels across $16$
scenarios. Each clip has a scenario tag, and scenario-masked decoding
restricts predictions to the allowed labels for that scenario.

\vspace{-1.0em}
\paragraph{Ego4D GoalStep.}
We use the Ego4D GoalStep actions \citep{song2024goalstep}: $9{,}137$
train segments from $480$ takes and $2{,}709$ evaluation segments from
$124$ take-disjoint validation takes. The closed label space has
$|\mathcal{V}_{\text{goalstep}}| = 310$ labels. Scenario masking is
enabled at decode time. We report Mid / Full $\macc$, $\sr$, mIoU, and
Levenshtein edit distance.

\vspace{-1.0em}
\paragraph{EgoPER.}
We use the EgoPER actions \citep{leeprocedural2024}: $2{,}018$ train
segments from $149$ videos and $921$ evaluation segments from $67$
take-disjoint validation videos, with $|\mathcal{V}_{\text{egoper}}| =
57$ labels over five recipe-style tasks. Scenario masking is disabled,
and the closed-vocabulary verifier uses unmasked Viterbi.

\subsection{Qwen3-VL endpoint cache (LLM-based baseline)}
\label{app:implementation:qwen-cache}

The Qwen3-VL SFT $+$ Qwen3-30B baseline first uses a supervised Qwen3-VL
endpoint-action cache to recognize the start and goal actions; Qwen3-30B
then receives those endpoint predictions plus the action-name bank and
fills the middle sequence. The endpoint cache is intentionally strong:
it is tuned on endpoint action recognition and reaches $43.58\%$
endpoint R@1 and $47.17\%$ endpoint R@5 over the unique start/goal
segments appearing in the $H{=}3$--$8$ planning manifests. The
LLM gap-filler is therefore not bottlenecked by raw visual parsing
alone; it tests whether accurate endpoint names plus a large LLM can
infer the missing procedure.

\subsection{Frame and trajectory preprocessing}
\label{app:preprocessing}

\paragraph{Visual frame preprocessing.}
We precompute one frozen V-JEPA~2.1 ViT-G/384 feature per action
segment and reuse this cache in all training and evaluation runs. The
predictor consumes a single $16$-frame action-centered clip per
segment. For \egoexo{} atomic actions, the action interval is derived
from the official atomic-action timestamps: the left and right
boundaries are the temporal midpoints to the neighboring
atomic-action timestamps within the same take. For benchmark
segments with their own timestamps, the interval is centered on the
annotated segment midpoint and clipped to the annotated segment
boundaries. Frames are resized to $384 \times 384$, normalized with
ImageNet mean and standard deviation, encoded by V-JEPA, and
mean-pooled into a single $1664$-dimensional segment feature.

\paragraph{Aria trajectory-control preprocessing.}
For \egoexo{} trajectory-conditioned experiments, the motion input
comes from the time-synchronized 6-DoF Aria-glasses pose trajectory
provided by \egoexo{}. Because the predictor expects a fixed-size
control tensor per action, we resample the Aria pose trajectory over
the corresponding action interval and express the sampled motion as
relative 6-DoF controls:
\[
[\Delta x,\Delta y,\Delta z,\Delta r_x,\Delta r_y,\Delta r_z].
\]
The first three values represent relative translation, and the last
three represent relative orientation in rotation-vector form.

\paragraph{PI3 trajectory preprocessing.}
For PI3 trajectory-source experiments, we replace the Aria pose
stream with RGB-camera poses estimated by PI3~\citep{pi3}. PI3
predicts a timestamped camera-pose sequence from $16$ RGB frames. We
align those poses to video time, interpolate translation and
rotation, and convert the result into the same relative 6-DoF control
format. PI3 controls match the predictor's input shape but come from
a different motion source: Aria controls are read off the glasses
pose stream, while PI3 controls are estimated from RGB appearance.
Aria 6-DoF poses are released only for \egoexo{}; for Keystep,
GoalStep, and EgoPER, the \oracletraj{} rows therefore use
PI3-estimated trajectories on the same ground-truth videos.

\subsection{VLM/LLM baseline details}
\label{app:vlm_baseline_details}

\paragraph{Qwen-ZS: zero-shot sequence generation.}
Qwen-ZS uses Qwen3-VL-32B without a LoRA adapter. For each horizon
$H \in \{3,\ldots,8\}$, the model receives 16 sampled RGB
frames per input segment, resized to a maximum side length of 336
pixels: start and goal segments for planning, the start segment only
for anticipation. For planning, the prompt is the following template,
with $H$ filled by the horizon and the full $8{,}472$-label action
bank appended as one action name per line:
\begin{quote}\small
You are given two ordered image groups from the same procedural
window. START endpoint frames show the first observed atomic action
and initial state. GOAL endpoint frames show the last observed atomic
action and final state. Predict exactly $H$ atomic actions from START
to GOAL. The first action should describe the START endpoint frames.
The last action should describe the GOAL endpoint frames. Fill any
intermediate actions that likely occur between them. Use only actions
from the action taxonomy when it is provided. Return JSON only. Use
this shape with exactly $H$ items:
\texttt{\{"sequence":[\{"action\_name":"..."\}, ...]\}}. Action
taxonomy: one valid action name per line.
\end{quote}
For anticipation, the goal frames are removed and the prompt becomes:
\begin{quote}\small
You are given one ordered image group from a procedural video. START
frames show the first observed atomic action and initial state. Predict
exactly $H$ atomic actions starting from this observed START action and
continuing with the most likely future actions. The first action should
describe the START frames. All remaining actions should be plausible
future atomic actions after the START action. No goal or final-state
frames are provided. Use only actions from the action taxonomy when it
is provided. Return JSON only. Use this shape with exactly $H$ items:
\texttt{\{"sequence":[\{"action\_name":"..."\}, ...]\}}. Action
taxonomy: one valid action name per line.
\end{quote}
The output is parsed as a JSON sequence and each generated action string
is matched against the $8{,}472$-label bank. Exact bank matches use the
matched label as top-1, with top-5 expanded by cosine nearest neighbors
over the frozen \egovlp{} bank embeddings. Strings that do not exactly
match a bank label are encoded with \egovlp{} and projected into the
same bank. Decoding is deterministic via vLLM with temperature $0.0$,
top-$p=1.0$, max output length $256$ tokens, max model length
$131{,}072$, and one sequence per prompt.

\paragraph{Endpoint-action recognizer (stage 1 of Qwen-SFT$+$LLM).}
The endpoint stage is a Qwen3-VL-32B LoRA SFT model trained on
$140{,}413$ atomic-segment records from the \egoexo{} atomic training
split. Each example contains 16 sampled segment frames at
$336$ pixel max side, with a target atomic action label. Each prompt
includes $32$ valid action-label exemplars sampled from different
scenarios, followed by the instruction
\begin{quote}\small
These time-ordered egocentric frames show one short atomic action.
Examples of valid atomic action label style: [32 example labels].
Identify the action being performed. Return only one concise lowercase
action label in the same style. Do not include JSON, IDs, punctuation,
explanations, or notes.
\end{quote}
SFT uses LoRA rank $8$, alpha $32$, dropout $0.05$, with target modules
\texttt{q\_proj}, \texttt{k\_proj}, \texttt{v\_proj}, \texttt{o\_proj},
\texttt{gate\_proj}, \texttt{up\_proj}, \texttt{down\_proj}. The base
model is loaded in bfloat16 with gradient checkpointing. Training runs
one epoch with AdamW, learning rate $2\times10^{-5}$, microbatch size
$1$, gradient accumulation $4$, on four H200 GPUs (seed $7$). Loss is
answer-only causal language modeling: prompt and padding tokens are
masked, so gradients flow only through the target action-label tokens.
Endpoint inference uses deterministic decoding (temperature $0.0$,
top-$p=1.0$, max output length $64$). The resulting cache covers
$10{,}849$ unique start/goal segments at $43.58\%$ R@1 and $47.17\%$
R@5 after bank projection (Table~\ref{tab:qwen3vl_endpoint_cache}).

\paragraph{LLM gap filling (stage 2 of Qwen-SFT$+$LLM).}
For each planning sample, the top-1 endpoint labels from stage 1
populate position $0$ and position $H{-}1$ of an incomplete length-$H$
sequence with \texttt{-1} placeholders for the missing middle. Qwen3-30B
receives the full $8{,}472$-label action bank, eight completed training
sequences from the same scenario when available, and the incomplete
sequence. The system prompt is:
\begin{quote}\small
You are a procedural planning assistant. Complete atomic action
sequences using only actions from the provided action bank. Return JSON
only. Do not write reasoning and do not output \texttt{<think>} tags.
\end{quote}
The user prompt specifies the sequence length, names the first and last
actions as fixed endpoint predictions, instructs the model to replace
only the intermediate \texttt{-1} placeholders, and requires
\texttt{\{"sequence":["<ACTION\_NAME>", ...]\}} with names matching the
bank exactly. Qwen thinking is disabled in the chat template, and any
residual thinking tags are stripped before parsing. Decoding is
deterministic (temperature $0.0$, top-$p=1.0$, max output length $256$,
one sequence per prompt) on the Transformers backend. Exact endpoint
IDs and exact generated bank strings are kept as top-1 with bank-neighbor
cosine expansion; non-exact strings are encoded with \egovlp{} and
projected into the same bank. This repaired $462$ unique unmapped LLM
strings and left a $0.00\%$ parse-failure rate at $H{=}3$--$8$.

\begin{table}[t]
\centering
\small
\setlength{\tabcolsep}{6pt}
\caption{Tuned Qwen3-VL-32B endpoint-action used as fixed start/goal
input to the LLM planner. R@5 is computed after projecting the generated
endpoint string into the atomic text bank and expanding with nearest
neighbors from the precomputed text-bank embeddings.}
\label{tab:qwen3vl_endpoint_cache}
\begin{tabular}{lrr}
\toprule
Quantity & Count & Percentage \\
\midrule
Endpoint segments & $10{,}849$ & -- \\
Endpoint R@1 & $4{,}728$ & $43.58$ \\
Endpoint R@5 & $5{,}118$ & $47.17$ \\
Exact text-bank projection + bank cosine & $10{,}838$ & $99.90$ \\
Fallback projection & $11$ & $0.10$ \\
\bottomrule
\end{tabular}
\end{table}
\subsection{Closed-vocabulary adaptation}
\label{app:implementation:closed-vocab}

For \egoexo{} Keystep, Ego4D GoalStep, and EgoPER, we adapt the same
trained backbone with a single benchmark-agnostic recipe. Three
properties of this recipe are critical to the transferability claim:
the alignment encoder $E_\tau$ is frozen across all three benchmarks
(no per-benchmark trajectory model is fit); a single shared
classifier-head template $\text{cls\_head}: \mathbb{R}^{768} \to
\mathbb{R}^{|\mathcal{V}|}$ is instantiated per benchmark with the
appropriate vocabulary size, while the backbone, attention pattern, and
goal-dropout schedule are unchanged; and the gate-then-rank scorer is
reused with only the readout swapped from the open-vocabulary
text-space scorer to a post-head sequence verifier that ranks
classifier-logit objects.

Per-benchmark hyperparameters
(Table~\ref{tab:closed_vocab_hparams}) vary in only a handful of
values: the predictor-loss weight $\lambda_{\text{bridge}}$, the
transition weight, the smoothing/temperature pair. All other architecture and optimization choices
are shared.

The scorer's retrieval pool is also reduced from
$K{=}64$ (atomic) to $K{=}5$ across all three closed-vocabulary
benchmarks: the smaller label space concentrates plausible
continuations on a few candidates, and a smaller pool simplifies
selection without losing recall in our diagnostics.

\begin{table}[t]
\centering
\scriptsize
\setlength{\tabcolsep}{4pt}
\caption{Per-benchmark hyperparameters for the closed-vocabulary
adaptation. All other architecture / optimization choices are shared
across benchmarks: Stage-1 \camformer{} frozen; same classifier-head
template; same Stage-2 backbone initialization.}
\label{tab:closed_vocab_hparams}
\begin{tabular}{lccccc}
\toprule
Benchmark & $|\mathcal{V}|$ & $\lambda_{\text{bridge}}$ & Trans. $w$ & Smooth / temp \\
\midrule
\egoexo{} Keystep & $375$ & $0.15$ & $1.5$  & $0.5 / 1.0$    \\
Ego4D GoalStep    & $310$ & $0.0$  & $0.05$ & $0.05 / 1.2$   \\
EgoPER            & $57$  & $0.0$  & $2.5$  & $0.5 / 1.0$  &  \\
\bottomrule
\end{tabular}
\end{table}

\subsection{Compute}
\label{app:compute}

All main TrajPilot training stages run on a single H200 GPU on an
internal cluster, with frozen V-JEPA~2.1 ViT-G features extracted
once and cached so that no encoder forward enters training cost.
Approximate wall-clock per stage: $30$ minutes for
trajectory--action alignment ($E_\tau$), $50$ minutes for the
open-vocabulary causal predictor on \egoexo{} atomic, $10$ minutes
per closed-vocabulary fine-tuning run (\egoexo{} Keystep, Ego4D
GoalStep, EgoPER), and $20$ minutes for the gate-then-rank scorer,
totaling roughly $2.2$ H200-hours of reported compute. The Qwen-SFT
endpoint cache additionally uses $4$ H200 GPUs for one epoch (see
\S\ref{app:implementation:qwen-cache}). Including preliminary and
ablation runs not appearing in the paper, the full project used
approximately $100\times$ the reported total ($\sim 220$ H200-hours).

\subsection{Assets and licenses}
\label{app:assets_licenses}

All datasets and pretrained models used in this paper are cited at
their canonical sources and used under their published research-use
terms. Datasets: Ego-Exo4D \citep{egoexo4d} (Ego-Exo4D Dataset License
Agreement, signed access); Ego4D / GoalStep \citep{song2024goalstep}
(Ego4D License Agreement, signed access); EgoPER
\citep{leeprocedural2024} (research use, distributed by author
request); EPIC-Kitchens-100 \citep{damen2022rescaling} (CC BY-NC 4.0).
Pretrained models: V-JEPA~2.1 \citep{vjepa21} (model weights Apache
2.0, code MIT); EgoVLPv2 \citep{lin2022egovlp} (MIT); CamFormer
\citep{xue2025seeing} (architecture reused; the trajectory encoder
$E_\tau$ is retrained from scratch in our pipeline); PI3 \citep{pi3}
(code 2-clause BSD; model weights for non-commercial research and
educational use only); Qwen3-VL-32B and Qwen3-30B (Apache 2.0).
Baseline planners SCHEMA, PDPP, and \viterbinet{} are re-implemented
from their published descriptions and retrained on matched \vjepag{}
features.

\section{Diagnostics}
\label{app:diagnostics}

\subsection{V-JEPA latent structure: scene dominates action}
\label{app:vjepa_geometry}

The diagnostic in \S\ref{sec:method:diagnostic} claims that V-JEPA
latents are organized by visual continuity rather than by action.
Table~\ref{tab:vjepa_geometry} quantifies this on Ego-Exo4D
atomic-action segments: clips from the same continuous recording are
substantially closer in V-JEPA space than clips of the same action
drawn from different recordings, and only random cross-recording
pairs are clearly far. The same ordering holds under both $\ell_1$
and cosine similarity.

\begin{table}[h]
\centering
\small
\caption{V-JEPA latent structure. Same-recording proximity exceeds
same-action cross-recording proximity, so raw latent distance is
not a clean action metric.}
\label{tab:vjepa_geometry}
\begin{tabular}{lcc}
\toprule
Pair type & Mean $\ell_1$ $\downarrow$ & Mean cosine $\uparrow$ \\
\midrule
Adjacent segments, same recording          & $0.171$ & $0.975$ \\
Far segments, same recording               & $0.205$ & $0.964$ \\
Same action text, different recording      & $0.256$ & $0.944$ \\
Random cross-recording pair                & $0.369$ & $0.886$ \\
\bottomrule
\end{tabular}
\end{table}

The consequence for CEM is direct. We measured the correlation
between V-JEPA latent $\ell_1$-to-goal and per-step action accuracy
across $2{,}000$ random training trajectories per horizon: Pearson
$-0.056$, $R^2 = 0.3\%$. Cosine variants explain only
$1$--$2\%$ of action variation. The objective CEM optimizes is
therefore close to uncorrelated with the metric we care about,
which is why search picks worse trajectories than no search at all.

\subsection{Goal-progress monotonicity collapses with horizon}
\label{app:monotonicity}

A second probe of the same problem: an $\ell_1$-to-goal scoring
objective implicitly assumes that real trajectories make
monotonic progress toward the goal in latent space.
Table~\ref{tab:monotonicity} reports, for $2{,}000$ random
training trajectories per horizon, the fraction whose latent
$\ell_1$ distance to the goal decreases monotonically as the plan
unfolds.

\begin{table}[h]
\centering
\small
\caption{Fraction of ground-truth trajectories that move
monotonically closer to the goal in V-JEPA latent $\ell_1$, by
horizon. Real human activity routinely moves away from the goal
before returning, and the assumption underlying $\ell_1$-to-goal
scoring breaks down beyond $H=3$.}
\label{tab:monotonicity}
\begin{tabular}{lcccccc}
\toprule
$H$ & 3 & 4 & 5 & 6 & 7 & 8 \\
\midrule
Monotonic fraction & $100\%$ & $51\%$ & $19\%$ & $6\%$ & $1.4\%$ & $0.1\%$ \\
\bottomrule
\end{tabular}
\end{table}

At $H=8$ only three of $2{,}000$ ground-truth trajectories satisfy
the assumption. The $\ell_1$-to-goal objective therefore penalizes
the trajectories CEM should be selecting.

\subsection{Latent-space CEM: full mean accuracy}
\label{app:cem_full_mean}

\begin{figure}[h]
\centering
\includegraphics[width=0.54\linewidth]{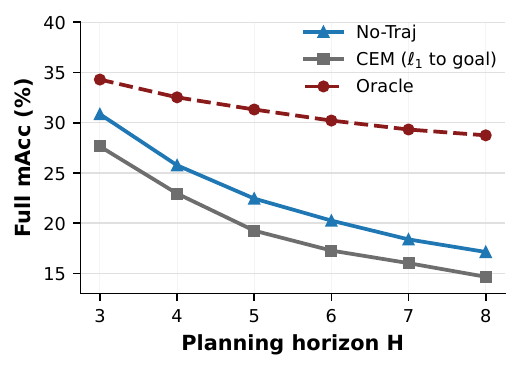}
\caption{Latent-space planning under Full mean accuracy
(companion to Figure~\ref{fig:cem_vs_none}). Same predictor and
three trajectory inputs.}
\label{fig:cem_full_mean}
\end{figure}

Figure~\ref{fig:cem_vs_none} in the main text reports Success Rate
for the No-Traj / CEM / Oracle comparison.
Figure~\ref{fig:cem_full_mean} reports the same comparison under
Full mean accuracy. The qualitative picture is identical: CEM
underperforms No-Traj at every horizon, while Oracle stays well
above both, with the gap widening as the horizon grows.

\subsection{Gate-then-rank scorer details}
\label{app:scorer_mechanism}

\begin{figure}[h]
\centering
\includegraphics[width=\linewidth]{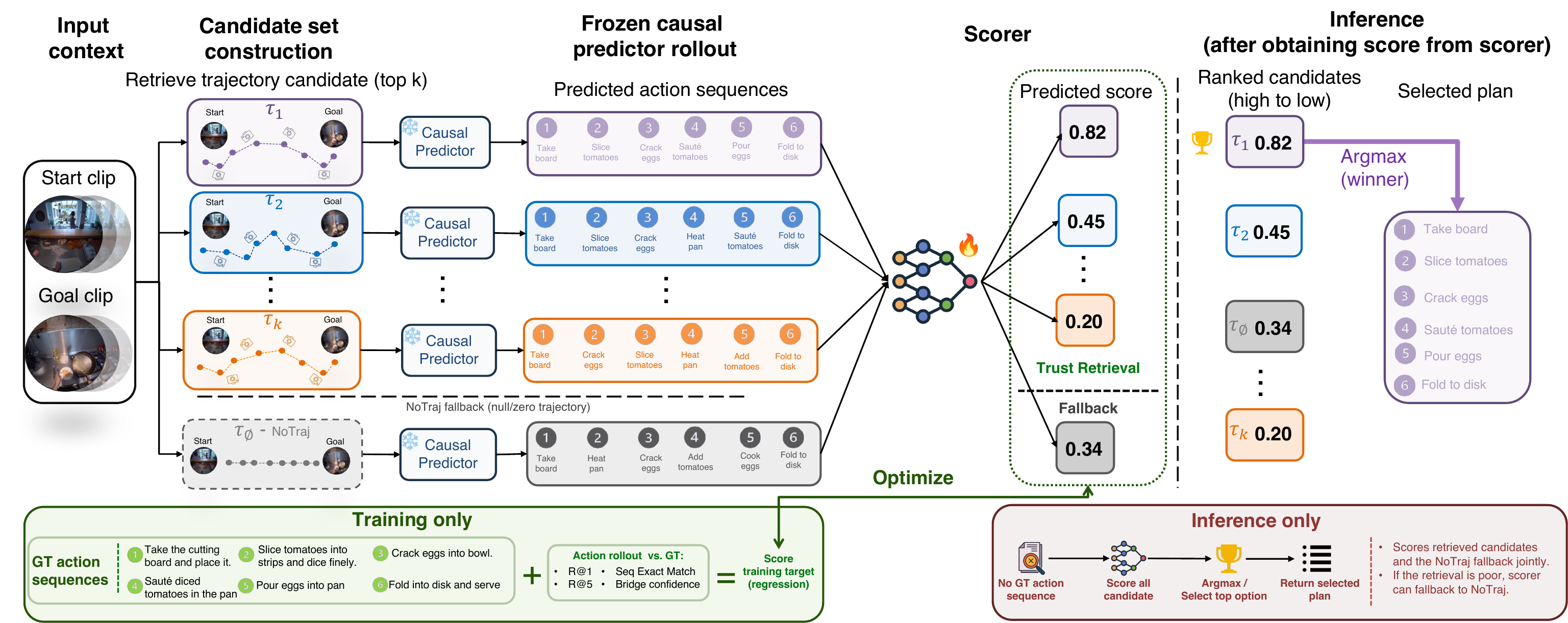}
\caption{Gate-then-rank scorer mechanism. Top-$K$ retrieved trajectory
candidates are rolled through the frozen trajectory-conditioned
predictor (left) to produce candidate plans; in parallel, the frozen
No-Traj predictor produces the fallback. The scorer (center) reads
all candidate plans plus the fallback and emits a binary gate logit
and per-candidate rank scores. At inference (right), the gate routes
to either the No-Traj fallback or the highest-rank retrieved
candidate. (Best viewed with zoom).}
\label{fig:scorer}
\end{figure}

Figure~\ref{fig:scorer} illustrates the mechanism of
\S\ref{sec:method:stage3}. The retrieval bank holds one entry per
training segment, indexed by Start/Goal endpoint embeddings. For each
test query, the top-$K{=}64$ bank entries are retrieved by endpoint
cosine and rolled through the frozen trajectory-conditioned predictor;
in parallel, the frozen No-Traj predictor produces the fallback. A
2-layer 4-head transformer scorer encodes, per candidate, the
candidate's predicted plan, the No-Traj prediction for the same
query, the retrieved trajectory's latent, the retrieved candidate's
action-label embedding, and a query vector from Start/Goal
V-JEPA features and trajectory-encoder endpoint latents. Two heads
emit (i) a binary gate logit and (ii) per-candidate rank scores.

\paragraph{Training objective.}\label{app:scorer_loss}
The scorer optimizes a retrieval-aligned utility,
\[
u_i \;=\; 0.1\, s_i^{\text{teacher}}
       \;+\; 1.0\, R@1_i
       \;+\; 0.5\, R@5_i
       \;+\; 1.0\, \mathrm{seq}_i,
\]
where $s_i^{\text{teacher}}$ is the frozen predictor's mean
ground-truth label logit and $R@1_i$, $R@5_i$, $\mathrm{seq}_i$ are
the candidate's atomic middle-step and exact-sequence retrieval
scores. The gate target is positive only when the best retrieved
candidate beats No-Traj on $u$ by a margin. On gate-positive
examples, the rank head is trained with cross-entropy to the best
candidate plus KL to the softmax over candidate utilities.

\paragraph{Selector behavior across horizons.}\label{app:selector_behavior}
The gate increasingly trusts retrieval as the horizon grows: it
fires ``No-Traj'' for roughly $81\%$ of $H{=}3$ test inputs and
$\sim 18\%$ at $H{=}8$. This tracks the Start--Goal uncertainty
cone: at long horizons many plans are consistent with the
endpoints, and the retrieved trajectory carries the discriminative
signal the gate elects to use.

\subsection{Retrieval headroom: the bottleneck is selection}
\label{app:retrieval_headroom}

The gate-then-rank scorer (\S\ref{sec:method:stage3}) ranks among $K=64$
candidates retrieved by Start/Goal cosine in the action-aligned
space. Table~\ref{tab:retrieval_headroom} shows that this pool is
rich enough to contain a relevant middle-step path for most test
inputs, so the inference problem is selecting from the pool rather
than expanding it.

\begin{table}[h]
\centering
\small
\caption{Endpoint retrieval headroom on Ego-Exo4D
at $H=5$. \emph{Same-step recall}: the
retrieved pool contains a trajectory whose mid-step at the matching
position is the ground-truth action. \emph{Any-step recall}: the
pool contains a trajectory whose mid-step at any position is the
ground-truth action. \emph{Cosine local/random}: mean Start/Goal
endpoint cosine within the retrieved pool versus a random pool.}
\label{tab:retrieval_headroom}
\begin{tabular}{lccc}
\toprule
Pool & Mid same-step recall & Mid any-step recall & Cosine local / random \\
\midrule
top-$1$  & $15.36\%$ & $25.33\%$ & $0.391 / 0.057$ \\
top-$5$  & $34.90\%$ & $46.09\%$ & $0.571 / 0.296$ \\
top-$16$ & $47.92\%$ & $57.81\%$ & $0.651 / 0.444$ \\
top-$64$ & $61.59\%$ & $73.31\%$ & $0.715 / 0.585$ \\
\bottomrule
\end{tabular}
\end{table}

To convert this recall into a planning upper bound, we replaced the
scorer with an oracle that picks the candidate from the top-$64$
pool whose plan best matches ground truth. At $H=5$ this oracle
reaches $42.2\%$ per-mid-step top-1 retrieval accuracy (M@1) and
$21.9\%$ exact mid-sequence match (MSeq), compared to No-Traj's
$20.5\%$ M@1 and $4.3\%$ MSeq. The retrieved pool therefore has
substantial headroom over the no-trajectory baseline; the role of
gate-then-rank is to recover as much of it as possible without
trajectory supervision at test time. By contrast, top-$1$
retrieval ($15.4\%$ same-step recall) underperforms No-Traj at
this horizon, confirming that naive nearest-neighbor lookup in the
trajectory bank is not enough.

\section{Full per-horizon results}
\label{app:full_results}

\begin{table*}[t]
\centering
\small
\setlength{\tabcolsep}{6pt}
\caption{\textbf{Open-vocabulary atomic planning, full per-horizon
results.} \egoexo{} atomic test split, $8{,}472$-label atomic
action bank. Methods are scored by mid-step retrieval (M@1, M@5),
mid-sequence exact-match (MSeq), and full-sequence counterparts (F@1,
F@5, FSeq) that include the observed Start and Goal. Entries are
percentages. Underlies Figure~\ref{fig:atomic_planning} in the main
text.}
\label{tab:atomic_results_full}
\begin{tabular}{ccrrrrrr}
\toprule
$H$ & Method & M@1 $\uparrow$ & M@5 $\uparrow$ & MSeq $\uparrow$ & F@1 $\uparrow$ & F@5 $\uparrow$ & FSeq $\uparrow$ \\
\midrule
\multirow{5}{*}{3}
 & Qwen-ZS                       & $2.86$  & $7.36$  & $2.86$  & $3.24$  & $8.53$  & $0.01$  \\
 & Qwen-SFT$+$LLM                & $10.87$ & $14.45$ & $10.87$ & $33.29$ & $36.88$ & $5.41$  \\
 & \ours{} (\notraj)             & $29.38$ & $45.59$ & $29.38$ & $31.11$ & $46.96$ & $10.20$ \\
 & \ours{} (\retrtraj)           & $\mathbf{35.24}$ & $\mathbf{45.75}$ & $\mathbf{35.24}$ & $\mathbf{33.21}$ & $\mathbf{46.57}$ & $\mathbf{11.90}$ \\
 & \ours{} (\oracletraj)         & $33.04$ & $43.93$ & $33.04$ & $33.31$ & $44.37$ & $12.91$ \\
\midrule
\multirow{5}{*}{4}
 & Qwen-ZS                       & $2.33$  & $7.25$  & $0.15$  & $2.83$  & $8.21$  & $0.00$  \\
 & Qwen-SFT$+$LLM                & $9.53$  & $11.83$ & $3.44$  & $27.19$ & $30.16$ & $1.89$  \\
 & \ours{} (\notraj)             & $24.20$ & $41.37$ & $11.00$ & $28.34$ & $44.91$ & $5.31$  \\
 & \ours{} (\retrtraj)           & $\mathbf{32.42}$ & $\mathbf{41.91}$ & $\mathbf{19.37}$ & $\mathbf{32.87}$ & $\mathbf{44.42}$ & $\mathbf{8.49}$  \\
 & \ours{} (\oracletraj)         & $33.55$ & $44.50$ & $19.50$ & $33.61$ & $44.68$ & $9.68$  \\
\midrule
\multirow{5}{*}{5}
 & Qwen-ZS                       & $1.99$  & $6.57$  & $0.01$  & $2.39$  & $7.54$  & $0.00$  \\
 & Qwen-SFT$+$LLM                & $9.54$  & $11.80$ & $1.38$  & $23.78$ & $26.57$ & $0.85$  \\
 & \ours{} (\notraj)             & $20.50$ & $37.43$ & $4.30$  & $25.37$ & $41.88$ & $2.60$  \\
 & \ours{} (\retrtraj)           & $\mathbf{29.79}$ & $\mathbf{37.64}$ & $\mathbf{11.88}$ & $\mathbf{31.56}$ & $\mathbf{41.24}$ & $\mathbf{6.00}$  \\
 & \ours{} (\oracletraj)         & $34.02$ & $44.74$ & $13.62$ & $33.98$ & $44.97$ & $7.39$  \\
\midrule
\multirow{5}{*}{6}
 & Qwen-ZS                       & $1.90$  & $6.58$  & $0.00$  & $2.25$  & $7.23$  & $0.00$  \\
 & Qwen-SFT$+$LLM                & $9.29$  & $11.52$ & $0.98$  & $21.28$ & $23.98$ & $0.62$  \\
 & \ours{} (\notraj)             & $18.09$ & $35.07$ & $1.95$  & $23.10$ & $39.63$ & $1.41$  \\
 & \ours{} (\retrtraj)           & $\mathbf{28.18}$ & $\mathbf{35.54}$ & $\mathbf{7.67}$  & $\mathbf{30.25}$ & $\mathbf{39.25}$ & $\mathbf{4.06}$  \\
 & \ours{} (\oracletraj)         & $34.10$ & $44.88$ & $10.32$ & $34.00$ & $45.01$ & $5.92$  \\
\midrule
\multirow{5}{*}{7}
 & Qwen-ZS                       & $1.62$  & $6.47$  & $0.00$  & $2.00$  & $7.18$  & $0.00$  \\
 & Qwen-SFT$+$LLM                & $9.76$  & $12.21$ & $0.15$  & $19.96$ & $22.76$ & $0.11$  \\
 & \ours{} (\notraj)             & $16.52$ & $32.89$ & $1.37$  & $21.26$ & $37.47$ & $1.13$  \\
 & \ours{} (\retrtraj)           & $\mathbf{25.89}$ & $\mathbf{33.53}$ & $\mathbf{5.14}$  & $\mathbf{28.40}$ & $\mathbf{37.34}$ & $\mathbf{3.05}$  \\
 & \ours{} (\oracletraj)         & $34.14$ & $45.00$ & $7.79$  & $34.11$ & $45.16$ & $4.55$  \\
\midrule
\multirow{5}{*}{8}
 & Qwen-ZS                       & $1.52$  & $6.47$  & $0.00$  & $1.85$  & $7.00$  & $0.00$  \\
 & Qwen-SFT$+$LLM                & $9.10$  & $11.35$ & $0.73$  & $18.23$ & $20.85$ & $0.41$  \\
 & \ours{} (\notraj)             & $15.11$ & $31.20$ & $1.20$  & $19.67$ & $35.74$ & $1.03$  \\
 & \ours{} (\retrtraj)           & $\mathbf{24.70}$ & $\mathbf{32.13}$ & $\mathbf{3.67}$  & $\mathbf{27.20}$ & $\mathbf{35.76}$ & $\mathbf{2.13}$  \\
 & \ours{} (\oracletraj)         & $34.39$ & $45.16$ & $6.12$  & $34.32$ & $45.28$ & $3.75$  \\
\bottomrule
\end{tabular}
\end{table*}

\begin{table*}[t]
\centering
\scriptsize
\setlength{\tabcolsep}{3pt}
\caption{\textbf{Closed-vocabulary procedural planning, full per-horizon
results across all four metrics.} Test sets: \egoexo{} Keystep
($|\mathcal{V}|{=}375$), Ego4D GoalStep ($|\mathcal{V}|{=}310$), and EgoPER ($|\mathcal{V}|{=}57$). All three share the \emph{same}
Stage-1\,+\,Stage-2 backbone, the same goal-dropout schedule, and the
same gate-then-rank machinery; only the per-benchmark hyperparameters
of Table~\ref{tab:closed_vocab_hparams} differ. SR is full-sequence
exact match; mAcc is mid-step mean accuracy; mIoU is full-sequence
IoU; ED is full-sequence Levenshtein edit distance. SR / mAcc / mIoU
are percentages, higher is better; ED is in steps, lower is better.
Best per (horizon, dataset, metric) cell is bolded. Underlies
Table~\ref{tab:coarse_planning_overall} in the main text.}
\label{tab:coarse_planning_full}
\begin{tabular}{ccrrrrrrrrrrrr}
\toprule
& & \multicolumn{4}{c}{Keystep ($|\mathcal{V}|{=}375$)} & \multicolumn{4}{c}{GoalStep ($|\mathcal{V}|{=}310$)} & \multicolumn{4}{c}{EgoPER ($|\mathcal{V}|{=}57$)} \\
\cmidrule(lr){3-6} \cmidrule(lr){7-10} \cmidrule(lr){11-14}
$H$ & Method & SR$\uparrow$ & mAcc$\uparrow$ & mIoU$\uparrow$ & ED$\downarrow$ & SR$\uparrow$ & mAcc$\uparrow$ & mIoU$\uparrow$ & ED$\downarrow$ & SR$\uparrow$ & mAcc$\uparrow$ & mIoU$\uparrow$ & ED$\downarrow$ \\
\midrule
\multirow{6}{*}{3}
 & SCHEMA                & $6.65$ & $17.42$ & $24.96$ & $2.29$ & $0.75$ & $4.55$ & $4.78$ & $2.86$ & $29.86$ & $43.84$ & $47.10$ & $1.69$ \\
 & PDPP                  & $4.97$ & $17.94$ & $23.22$ & $2.29$ & $0.50$ & $4.84$ & $7.20$ & $2.78$ & $\mathbf{84.75}$ & $\mathbf{87.80}$ & $91.39$ & $\mathbf{0.30}$ \\
 & \viterbinet{}         & $\mathbf{8.97}$ & $20.72$ & $28.72$ & $2.14$ & $1.00$ & $7.51$ & $8.15$ & $2.77$ & $67.22$ & $78.65$ & $82.88$ & $0.61$ \\
 & \ours{} (\notraj)     & $7.89$ & $20.15$ & $28.86$ & $2.13$ & $\mathbf{1.38}$ & $7.01$ & $10.77$ & $\mathbf{2.65}$ & $83.86$ & $87.17$ & $91.33$ & $0.32$ \\
 & \ours{} (\retrtraj)   & $8.07$ & $20.70$ & $\mathbf{29.43}$ & $\mathbf{2.10}$ & $1.34$ & $6.72$ & $10.82$ & $2.65$ & $84.24$ & $87.29$ & $\mathbf{91.48}$ & $0.31$ \\
 & \ours{} (\oracletraj) & $8.69$ & $\mathbf{22.63}$ & $29.42$ & $2.11$ & $1.21$ & $\mathbf{7.68}$ & $\mathbf{10.96}$ & $2.65$ & $80.81$ & $84.75$ & $89.69$ & $0.39$ \\
\midrule
\multirow{6}{*}{4}
 & SCHEMA                & $3.35$ & $14.95$ & $23.82$ & $3.16$ & $0.57$ & $3.45$ & $3.24$ & $3.86$ & $27.92$ & $49.79$ & $52.40$ & $2.02$ \\
 & PDPP                  & $2.87$ & $16.13$ & $23.87$ & $3.09$ & $0.40$ & $5.33$ & $9.22$ & $3.67$ & $78.33$ & $85.14$ & $90.10$ & $0.48$ \\
 & \viterbinet{}         & $\mathbf{4.27}$ & $17.29$ & $27.55$ & $2.99$ & $\mathbf{0.93}$ & $7.56$ & $10.19$ & $3.59$ & $58.61$ & $77.85$ & $81.46$ & $0.86$ \\
 & \ours{} (\notraj)     & $3.24$ & $17.91$ & $29.00$ & $2.91$ & $0.66$ & $6.61$ & $9.67$ & $3.61$ & $79.44$ & $86.74$ & $91.90$ & $\mathbf{0.44}$ \\
 & \ours{} (\retrtraj)   & $3.38$ & $18.06$ & $29.20$ & $2.90$ & $0.88$ & $7.16$ & $10.56$ & $3.58$ & $\mathbf{79.58}$ & $86.74$ & $\mathbf{91.99}$ & $0.43$ \\
 & \ours{} (\oracletraj) & $4.16$ & $\mathbf{21.24}$ & $\mathbf{29.52}$ & $\mathbf{2.87}$ & $0.80$ & $\mathbf{7.87}$ & $\mathbf{10.77}$ & $\mathbf{3.57}$ & $78.47$ & $85.28$ & $89.90$ & $0.48$ \\
\midrule
\multirow{6}{*}{5}
 & SCHEMA                & $1.56$ & $12.41$ & $23.26$ & $4.08$ & $0.51$ & $5.92$ & $6.08$ & $4.69$ & $24.81$ & $46.96$ & $56.04$ & $2.48$ \\
 & PDPP                  & $1.28$ & $14.18$ & $23.09$ & $3.99$ & $0.19$ & $3.36$ & $6.63$ & $4.73$ & $69.83$ & $81.01$ & $89.21$ & $0.73$ \\
 & \viterbinet{}         & $\mathbf{2.52}$ & $15.38$ & $28.32$ & $3.82$ & $\mathbf{0.79}$ & $7.33$ & $10.76$ & $4.52$ & $51.15$ & $72.18$ & $80.66$ & $1.20$ \\
 & \ours{} (\notraj)     & $2.27$ & $16.79$ & $29.67$ & $3.70$ & $0.61$ & $7.33$ & $10.48$ & $4.50$ & $71.52$ & $82.29$ & $\mathbf{91.18}$ & $0.66$ \\
 & \ours{} (\retrtraj)   & $2.33$ & $16.89$ & $\mathbf{29.70}$ & $3.69$ & $0.75$ & $7.08$ & $10.70$ & $4.50$ & $72.28$ & $82.90$ & $\mathbf{91.42}$ & $0.64$ \\
 & \ours{} (\oracletraj) & $2.41$ & $\mathbf{20.70}$ & $29.52$ & $\mathbf{3.63}$ & $\mathbf{0.79}$ & $\mathbf{7.97}$ & $\mathbf{10.93}$ & $\mathbf{4.49}$ & $\mathbf{74.73}$ & $\mathbf{84.48}$ & $90.42$ & $\mathbf{0.60}$ \\
\midrule
\multirow{6}{*}{6}
 & SCHEMA                & $1.01$ & $12.13$ & $23.26$ & $4.90$ & $0.49$ & $6.15$ & $7.23$ & $5.60$ & $22.70$ & $45.95$ & $62.06$ & $2.72$ \\
 & PDPP                  & $0.69$ & $13.54$ & $23.80$ & $4.84$ & $0.00$ & $4.32$ & $8.32$ & $5.62$ & $65.02$ & $79.39$ & $88.69$ & $0.95$ \\
 & \viterbinet{}         & $\mathbf{1.49}$ & $14.61$ & $28.31$ & $4.60$ & $0.49$ & $7.56$ & $8.76$ & $5.50$ & $48.12$ & $72.57$ & $82.55$ & $1.38$ \\
 & \ours{} (\notraj)     & $1.01$ & $15.58$ & $\mathbf{29.63}$ & $4.52$ & $\mathbf{0.69}$ & $6.73$ & $9.92$ & $5.45$ & $63.99$ & $80.03$ & $89.84$ & $0.92$ \\
 & \ours{} (\retrtraj)   & $0.95$ & $15.55$ & $29.57$ & $4.52$ & $0.69$ & $6.45$ & $10.00$ & $5.46$ & $65.19$ & $80.63$ & $89.84$ & $0.90$ \\
 & \ours{} (\oracletraj) & $1.25$ & $\mathbf{19.05}$ & $29.05$ & $\mathbf{4.45}$ & $0.64$ & $\mathbf{7.12}$ & $\mathbf{10.85}$ & $\mathbf{5.44}$ & $\mathbf{70.48}$ & $\mathbf{83.87}$ & $\mathbf{90.14}$ & $\mathbf{0.77}$ \\
\midrule
\multirow{6}{*}{7}
 & SCHEMA                & $0.41$ & $12.08$ & $24.17$ & $5.71$ & $0.36$ & $5.46$ & $6.48$ & $6.58$ & $18.30$ & $42.35$ & $57.89$ & $3.35$ \\
 & PDPP                  & $0.47$ & $14.54$ & $26.16$ & $5.50$ & $0.16$ & $5.10$ & $9.55$ & $6.51$ & $61.66$ & $77.92$ & $87.57$ & $1.19$ \\
 & \viterbinet{}         & $0.47$ & $11.55$ & $23.39$ & $5.75$ & $0.31$ & $6.66$ & $10.45$ & $6.42$ & $47.01$ & $70.40$ & $80.61$ & $1.74$ \\
 & \ours{} (\notraj)     & $0.72$ & $14.86$ & $\mathbf{29.93}$ & $5.31$ & $0.36$ & $6.83$ & $10.18$ & $6.37$ & $62.04$ & $79.96$ & $\mathbf{89.98}$ & $1.05$ \\
 & \ours{} (\retrtraj)   & $\mathbf{0.88}$ & $14.84$ & $29.59$ & $5.31$ & $\mathbf{0.68}$ & $7.25$ & $10.85$ & $6.34$ & $62.04$ & $79.96$ & $89.98$ & $1.05$ \\
 & \ours{} (\oracletraj) & $0.82$ & $\mathbf{18.79}$ & $29.43$ & $\mathbf{5.22}$ & $0.68$ & $\mathbf{8.07}$ & $\mathbf{11.47}$ & $\mathbf{6.30}$ & $\mathbf{66.09}$ & $\mathbf{82.97}$ & $89.72$ & $\mathbf{0.93}$ \\
\midrule
\multirow{6}{*}{8}
 & SCHEMA                & $0.17$ & $11.35$ & $23.66$ & $6.56$ & $0.22$ & $5.47$ & $5.13$ & $7.56$ & $19.69$ & $42.70$ & $60.52$ & $3.66$ \\
 & PDPP                  & $0.20$ & $11.03$ & $23.52$ & $6.59$ & $0.00$ & $6.47$ & $10.48$ & $7.39$ & $\mathbf{62.83}$ & $77.65$ & $86.88$ & $1.41$ \\
 & \viterbinet{}         & $0.13$ & $11.99$ & $24.41$ & $6.50$ & $0.27$ & $6.85$ & $10.31$ & $7.34$ & $46.46$ & $68.95$ & $77.65$ & $2.15$ \\
 & \ours{} (\notraj)     & $0.20$ & $14.50$ & $\mathbf{30.20}$ & $\mathbf{6.06}$ & $0.27$ & $6.23$ & $9.70$ & $7.36$ & $60.18$ & $79.17$ & $89.85$ & $1.24$ \\
 & \ours{} (\retrtraj)   & $0.23$ & $14.46$ & $30.01$ & $6.07$ & $\mathbf{0.55}$ & $6.50$ & $10.85$ & $7.30$ & $60.18$ & $79.17$ & $89.85$ & $1.24$ \\
 & \ours{} (\oracletraj) & $\mathbf{0.53}$ & $\mathbf{18.26}$ & $30.15$ & $5.99$ & $0.49$ & $\mathbf{7.92}$ & $\mathbf{11.87}$ & $\mathbf{7.22}$ & $63.50$ & $\mathbf{82.89}$ & $\mathbf{89.94}$ & $\mathbf{1.02}$ \\
\bottomrule
\end{tabular}
\end{table*}

\begin{table*}[t]
\centering
\small
\setlength{\tabcolsep}{3pt}
\caption{\textbf{Open-vocabulary atomic action anticipation, full
per-horizon results.} \egoexo{} atomic test split,
$8{,}472$-label atomic action bank, goal removed at inference. \ours{}
uses the same Stage-2 checkpoint as planning evaluated with
\texttt{anticipation\_mode}\,$=$\,\texttt{True}; \retrtraj{} uses the
start-only top-$64$ scorer. VLM baselines generate sequences from the
start clip only and project into the same action bank. Future R@1 /
R@5 / Seq are computed over the predicted future steps; Full R@1 / R@5
/ Seq additionally include the observed Start. Underlies
Figure~\ref{fig:atomic_anticipation} in the main text.}
\label{tab:atomic_anticipation_full}
\begin{tabular}{ccrrrrrr}
\toprule
$H$ & Method & Future R@1 $\uparrow$ & Future R@5 $\uparrow$ & Future Seq $\uparrow$ & Full R@1 $\uparrow$ & Full R@5 $\uparrow$ & Full Seq $\uparrow$ \\
\midrule
\multirow{5}{*}{3}
 & Qwen-ZS                       & $1.22$  & $7.24$  & $1.22$  & $1.78$  & $7.41$  & $0.01$  \\
 & Qwen-SFT$+$LLM                & $11.79$ & $14.57$ & $4.27$  & $22.68$ & $25.71$ & $3.08$  \\
 & \ours{} (\notraj)             & $26.20$ & $38.43$ & $26.20$ & $24.98$ & $37.13$ & $\mathbf{5.13}$  \\
 & \ours{} (\retrtraj)           & $30.56$ & $38.65$ & $30.56$ & $26.55$ & $36.90$ & $5.71$  \\
 & \ours{} (\oracletraj)         & $\mathbf{33.91}$ & $\mathbf{44.31}$ & $\mathbf{33.91}$ & $\mathbf{26.84}$ & $\mathbf{37.32}$ & $5.48$  \\
\midrule
\multirow{5}{*}{4}
 & Qwen-ZS                       & $1.07$  & $6.11$  & $0.07$  & $1.67$  & $6.95$  & $0.00$  \\
 & Qwen-SFT$+$LLM                & $10.85$ & $13.40$ & $1.31$  & $19.32$ & $22.11$ & $1.06$  \\
 & \ours{} (\notraj)             & $21.34$ & $35.33$ & $9.33$  & $22.94$ & $35.83$ & $2.96$  \\
 & \ours{} (\retrtraj)           & $25.93$ & $33.56$ & $13.59$ & $25.15$ & $34.54$ & $3.44$  \\
 & \ours{} (\oracletraj)         & $\mathbf{33.86}$ & $\mathbf{44.43}$ & $\mathbf{19.83}$ & $\mathbf{28.66}$ & $\mathbf{39.19}$ & $\mathbf{4.69}$  \\
\midrule
\multirow{5}{*}{5}
 & Qwen-ZS                       & $1.38$  & $5.89$  & $0.02$  & $1.59$  & $7.19$  & $0.00$  \\
 & Qwen-SFT$+$LLM                & $10.96$ & $13.86$ & $0.68$  & $17.73$ & $20.75$ & $0.57$  \\
 & \ours{} (\notraj)             & $18.18$ & $32.79$ & $3.74$  & $20.46$ & $33.98$ & $1.78$  \\
 & \ours{} (\retrtraj)           & $23.73$ & $30.79$ & $7.32$  & $23.59$ & $32.25$ & $2.08$  \\
 & \ours{} (\oracletraj)         & $\mathbf{33.89}$ & $\mathbf{44.46}$ & $\mathbf{13.48}$ & $\mathbf{29.63}$ & $\mathbf{40.13}$ & $\mathbf{3.34}$  \\
\midrule
\multirow{5}{*}{6}
 & Qwen-ZS                       & $1.34$  & $6.50$  & $0.00$  & $1.48$  & $6.99$  & $0.00$  \\
 & Qwen-SFT$+$LLM                & $10.44$ & $13.02$ & $0.34$  & $16.17$ & $18.91$ & $0.29$  \\
 & \ours{} (\notraj)             & $16.37$ & $31.56$ & $2.15$  & $19.13$ & $33.14$ & $1.61$  \\
 & \ours{} (\retrtraj)           & $22.18$ & $28.77$ & $4.85$  & $22.69$ & $30.70$ & $1.96$  \\
 & \ours{} (\oracletraj)         & $\mathbf{33.84}$ & $\mathbf{44.35}$ & $\mathbf{9.70}$  & $\mathbf{30.39}$ & $\mathbf{40.91}$ & $\mathbf{2.78}$  \\
\midrule
\multirow{5}{*}{7}
 & Qwen-ZS                       & $1.29$  & $6.50$  & $0.00$  & $1.54$  & $7.30$  & $0.00$  \\
 & Qwen-SFT$+$LLM                & $10.39$ & $13.20$ & $0.17$  & $15.32$ & $18.24$ & $0.13$  \\
 & \ours{} (\notraj)             & $15.26$ & $30.64$ & $1.77$  & $17.79$ & $32.18$ & $1.15$  \\
 & \ours{} (\retrtraj)           & $21.23$ & $27.86$ & $3.22$  & $21.91$ & $29.69$ & $1.39$  \\
 & \ours{} (\oracletraj)         & $\mathbf{33.98}$ & $\mathbf{44.49}$ & $\mathbf{7.33}$  & $\mathbf{31.03}$ & $\mathbf{41.50}$ & $\mathbf{2.37}$  \\
\midrule
\multirow{5}{*}{8}
 & Qwen-ZS                       & $1.39$  & $6.63$  & $0.00$  & $1.52$  & $7.15$  & $0.00$  \\
 & Qwen-SFT$+$LLM                & $11.12$ & $14.16$ & $0.21$  & $15.36$ & $18.47$ & $0.18$  \\
 & \ours{} (\notraj)             & $14.22$ & $29.79$ & $1.53$  & $16.80$ & $31.46$ & $1.25$  \\
 & \ours{} (\retrtraj)           & $20.09$ & $27.19$ & $2.35$  & $21.00$ & $29.04$ & $1.36$  \\
 & \ours{} (\oracletraj)         & $\mathbf{33.87}$ & $\mathbf{44.51}$ & $\mathbf{5.39}$  & $\mathbf{31.30}$ & $\mathbf{41.94}$ & $\mathbf{1.91}$  \\
\bottomrule
\end{tabular}
\end{table*}

\subsection{Goal dropout}
\label{app:goal_dropout_full}

The body (\S\ref{sec:exp:anticipation}) reports a single
sample-weighted-overall Future R@1 number for the anticipation gain
from goal dropout. Tables~\ref{tab:goal_dropout_fullgoal} and
\ref{tab:goal_dropout_anticipation_full} give the full per-horizon
detail underneath that summary, with Future R@1 / Future R@5 / Future
Seq in each cell.
Table~\ref{tab:goal_dropout_fullgoal} supports the no-harm claim made
in the body for standard full-goal planning;
Table~\ref{tab:goal_dropout_anticipation_full} gives the per-horizon
view of the anticipation gain summarized in
\S\ref{sec:exp:anticipation}.

\begin{table*}[t]
\centering
\small
\setlength{\tabcolsep}{4pt}
\caption{%
Goal-dropout comparison under the standard full-goal planning
setting. Traj rows receive full goal context plus trajectory conditioning;
No-Traj rows receive full goal context without trajectory conditioning.
Entries are Future R@1 / Future R@5 / Future Seq.}
\label{tab:goal_dropout_fullgoal}
\begin{tabular}{ccccc}
\toprule
$H$ & No-dropout Traj & Goal-dropout Traj & No-dropout No-Traj & Goal-dropout No-Traj \\
\midrule
3 & $33.0 / 43.9 / 33.0$ & $34.5 / 45.6 / 34.5$ & $29.1 / 45.1 / 29.1$ & $31.6 / 45.6 / 31.6$ \\
4 & $33.6 / 44.5 / 19.5$ & $34.6 / 45.5 / 20.2$ & $24.0 / 41.0 / 10.9$ & $26.7 / 41.3 / 13.6$ \\
5 & $34.0 / 44.7 / 13.6$ & $34.7 / 45.6 / 13.6$ & $20.3 / 37.1 / 4.3$ & $22.9 / 38.4 / 5.8$ \\
6 & $34.1 / 44.9 / 10.3$ & $34.6 / 45.7 / 9.9$  & $17.9 / 34.7 / 1.9$ & $20.2 / 36.2 / 2.6$ \\
7 & $34.1 / 45.0 / 7.8$  & $34.7 / 45.7 / 7.5$  & $16.4 / 32.7 / 1.4$ & $18.4 / 35.0 / 2.0$ \\
8 & $34.4 / 45.2 / 6.1$  & $34.7 / 45.7 / 5.7$  & $15.0 / 31.0 / 1.2$ & $17.1 / 34.0 / 1.4$ \\
\midrule
Overall & $33.8 / 44.7 / 15.8$ & $34.6 / 45.6 / 16.0$ & $20.8 / 37.3 / 8.8$ & $23.2 / 38.7 / 10.3$ \\
\bottomrule
\end{tabular}
\end{table*}

\vspace{-1.0em}
\paragraph{Trade-off at long horizons.} Goal dropout slightly hurts
exact-sequence match (Future Seq) in the full-goal Traj setting at
long horizons: at most $-0.4$ pp at $H{=}8$ (Future Seq
$6.1 \to 5.7$). The most plausible reading is that masking the goal
half the time during training regularizes the model toward goal-free
behavior, trading a small amount of peak full-goal exact-match for
robustness to a missing goal. The cost is uniformly small (under
$1$ pp on every (horizon, metric) cell of
Table~\ref{tab:goal_dropout_fullgoal}), while the corresponding
anticipation gain is large ($+7.5$ pp Future R@1 overall without
trajectory; \S\ref{sec:exp:anticipation}), so we use the
goal-dropout checkpoint for anticipation and the no-dropout
checkpoint for planning.

\begin{table*}[t]
\centering
\small
\setlength{\tabcolsep}{4pt}
\caption{Goal-dropout comparison under goal-free action anticipation.
No-dropout columns zero the goal representation only at inference;
goal-dropout columns are trained with per-sample goal dropout and
evaluated with the same goal-free input. Entries are
Future R@1 / Future R@5 / Future Seq.}
\label{tab:goal_dropout_anticipation_full}
\begin{tabular}{ccccc}
\toprule
$H$ & No-dropout Traj & Goal-dropout Traj & No-dropout No-Traj & Goal-dropout No-Traj \\
\midrule
3 & $29.4 / 39.7 / 29.4$ & $33.9 / 44.3 / 33.9$ & $14.7 / 25.6 / 14.7$ & $26.0 / 38.2 / 26.0$ \\
4 & $30.3 / 40.3 / 16.2$ & $33.9 / 44.4 / 19.8$ & $12.9 / 23.9 / 3.4$  & $21.1 / 35.0 / 9.2$ \\
5 & $30.9 / 40.8 / 11.0$ & $33.9 / 44.5 / 13.5$ & $11.1 / 21.7 / 1.5$  & $18.0 / 32.5 / 3.7$ \\
6 & $31.1 / 41.0 / 7.7$  & $33.8 / 44.3 / 9.7$  & $9.7 / 20.3 / 1.3$   & $16.2 / 31.3 / 2.1$ \\
7 & $31.3 / 41.3 / 5.7$  & $34.0 / 44.5 / 7.3$  & $9.3 / 19.7 / 1.1$   & $15.1 / 30.4 / 1.8$ \\
8 & $31.5 / 41.7 / 4.2$  & $33.9 / 44.5 / 5.4$  & $8.5 / 19.1 / 1.0$   & $14.1 / 29.6 / 1.5$ \\
\midrule
Overall & $30.7 / 40.8 / 13.0$ & $33.9 / 44.4 / 15.7$ & $11.2 / 21.9 / 4.1$ & $18.7 / 33.1 / 8.0$ \\
\bottomrule
\end{tabular}
\end{table*}

\section{EPIC-Kitchens-100 cross-corpus transfer}
\label{app:ek100}

We test cross-corpus transfer of \ours{} to EPIC-Kitchens-100
(EK100) action anticipation~\citep{damen2022rescaling}, building on
the official V-JEPA~2 $4$-second full-token attentive-probe
protocol~\citep{vjepa21}. The visual encoder is kept frozen.
Because the V-JEPA~2.1 attentive-probe checkpoint is not publicly
released at submission time, we use a hybrid setup: the attentive
probe consumes full-token V-JEPA~2 features (the released checkpoint
finetuned on EK100), while our predictor consumes mean-pooled
V-JEPA~2.1 ViT-g (1B) features~\citep{assran2025vjepa2}.
We adapt the predictor to EK100 so that it maps these features
from the $[t{-}2, t{-}1]$ pre-action window to an action-pair
embedding in the EK100 text space rather than the open-vocabulary
atomic bank; the predicted embedding is injected as one additional
token into the attentive probe. The
gate-then-rank scorer is unchanged: at test time it ranks among
top-$5$ retrieved candidate trajectories from the EK100 training
set, with a no-trajectory fallback. Results in Table~\ref{tab:ek100}.

\begin{table}[t]
\centering
\small
\setlength{\tabcolsep}{4pt}
\caption{EPIC-Kitchens-100 anticipation, EK100-val. Verb / Noun /
Action are mean-class Recall@5 (\%) on the official $3{,}568$
action-pair label space; Action Acc.\ is instance accuracy (not
reported in the literature). \emph{Top:} numbers reported in prior
publications. \emph{Bottom:} our run, with V-JEPA~2.1 ViT-g features
and the V-JEPA~2 attentive-probe checkpoint, finetuned on EK100.
\ours{} variants inject the predictor's predicted action-pair
embedding as one additional token into the probe. Bolding marks the
best deployable method per column. The hybrid V-JEPA~2.1 features $+$
V-JEPA~2 probe setup is forced by checkpoint availability and
explains why our visual-only row sits below both V-JEPA~2 ViT-g
($39.7$) and V-JEPA~2.1 ViT-g ($38.4$) published Action R@5.}
\label{tab:ek100}
\begin{tabular}{lccccc}
\toprule
Method & Params & Verb & Noun & Action & Action Acc. \\
\midrule
\multicolumn{6}{l}{\emph{Reported in the literature}} \\
InAViT~\citep{roy2024inavit}                       & 160M & $51.9$ & $52.0$ & $25.8$ & --- \\
Video-LLaMA~\citep{zhang2023videollama}            & 7B   & $52.9$ & $52.0$ & $26.0$ & --- \\
PlausiVL~\citep{mittal2024plausivl}                & 8B   & $55.6$ & $54.2$ & $27.6$ & --- \\
V-JEPA~2 ViT-g~\citep{assran2025vjepa2}            & 1B   & $63.6$ & $57.1$ & $39.7$ & --- \\
V-JEPA~2.1 ViT-g~\citep{vjepa21}                   & 1B   & $63.6$ & $56.2$ & $38.4$ & --- \\
V-JEPA~2.1 ViT-G~\citep{vjepa21}                   & 2B   & $64.3$ & $59.9$ & $40.8$ & --- \\
\midrule
\multicolumn{6}{l}{\emph{Our run: V-JEPA~2.1 features $+$ V-JEPA~2 probe checkpoint, finetuned}} \\
Visual-only attentive probe                        & 1B   & $60.21$ & $56.39$ & $37.77$ & $64.07$ \\
\ours{} (\notraj{})                                & 1B   & $60.62$ & $56.70$ & $\mathbf{38.27}$ & $\mathbf{65.42}$ \\
\ours{} (\retrtraj{})                              & 1B   & $\mathbf{60.74}$ & $\mathbf{57.23}$ & $38.21$ & $65.41$ \\
\midrule
\ours{} (\oracletraj{})                            & 1B   & $61.79$ & $57.59$ & $38.55$ & $65.95$ \\
\bottomrule
\end{tabular}
\end{table}

In the matched ViT-g (1B) hybrid setting, the predictor's
action-aligned readout token (\ours{} \notraj{}) lifts every metric
over the visual-only probe, with the largest gain on instance Action
accuracy ($+1.35$ pp). Adding retrieved trajectory (\ours{} \retrtraj{})
gives a further small bump on Verb and Noun mean-class R@5 but is a
wash on the joint Action metrics; even ground-truth trajectory
(\ours{} \oracletraj{}) adds only $+0.78$ pp Action R@5 over the
visual-only probe. EK100 anticipation is dominated by long-tail
recognition rather than near-future disambiguation, so trajectory
headroom on this corpus is small. The value of this transfer
experiment is showing that the recipe applies to a different corpus
and protocol without retraining the alignment encoder, with the
action-aligned readout doing most of the work.

\section{Basketball shot-outcome prediction}
\label{app:basketball}

To test the event-prediction instantiation of the framework
(\S\ref{sec:method:setup}), we apply \ours{} to predicting whether a
basketball shot will score from pre-shot egocentric context on
Ego-Exo4D~\citep{egoexo4d}. The task is binary (made vs.\ missed);
the model never sees the release or result frames, only the context
that precedes the shot. Here $H{=}2$ denotes one pre-shot context
segment plus one predicted future shot-action token, distinct from
the planning horizon $H$ used elsewhere.

\paragraph{Data.}
We mine a binary shot-outcome split directly from
Ego-Exo4D~\citep{egoexo4d} using its own annotations, without any
hand selection. We extract every segment whose action caption matches
a shot-attempt pattern (\textit{shoot hoop}, \textit{shoot jump\_shot},
\textit{shoot layup}, \textit{attempt layup}), then derive the binary
outcome from the post-shot annotation text: a clip is labelled
\emph{made} if the next caption contains language consistent with the
ball entering the basket (e.g.\ ``shoots a jump shot into the hoop'',
``enters the net'') and \emph{missed} if it contains miss evidence
(e.g.\ ``bounces off the rim'', ``hits the backboard'',
``rebound ball'', ``miss hoop''). This procedure yields $376$ shot
events across $52$ basketball takes. We then class-balance the set to
$184$ made / $184$ missed and split take-disjointly into $294$ train
events ($147 / 147$, $42$ takes) and $74$ test events ($37 / 37$,
$10$ takes). The context fed to every model is strictly the pre-shot
preparation segment ($1.0$\,s on average; $16$ uniformly sampled
frames at the source frame rate of $30$\,fps); no model sees the
release or post-release frames, so the outcome can be inferred only
from the body's setup motion.

\paragraph{Setup.}
The alignment encoder $E_\tau$ stays at its atomic-pretrained
checkpoint; the causal predictor is fine-tuned with a binary
classification head on mean-pooled output tokens (AdamW, $80$ epochs,
batch $64$, lr $3{\times}10^{-4}$, weight decay $10^{-4}$). The
scorer ranks among top-$32$ retrieved candidate trajectories from the
basketball train set, with a no-trajectory fallback. We compare four
settings (Table~\ref{tab:basketball_shot_outcome}).
\textbf{\ours{} (\notraj)} receives only the pre-shot context.
\textbf{V-JEPA~2.1 attentive probe} uses the same context but
classifies from full V-JEPA~2.1 visual tokens ($4{,}608$ context plus
$576$ predicted future tokens) via an attentive outcome head.
\textbf{\ours{} (\retrtraj)} adds retrieved future trajectories with
the scorer selecting between fallback and retrieved
candidate. \textbf{\ours{} (\oracletraj)} feeds the predictor the
ground-truth future trajectory of the upcoming shot as an upper
bound.

\begin{table}[t]
\centering
\small
\setlength{\tabcolsep}{6pt}
\caption{Basketball shot-outcome prediction on a balanced
take-disjoint split mined from Ego-Exo4D ($74$ test shots, $37$ made
/ $37$ missed). Outcome Acc.\ is overall accuracy; Balanced Acc.\ is
the mean of made and missed recall (equal under a balanced split).}
\label{tab:basketball_shot_outcome}
\begin{tabular}{lcccc}
\toprule
Method & Outcome Acc. & Balanced Acc. & Made Rec. & Miss Rec. \\
\midrule
\ours{} (\notraj{})                       & $56.76$ & $56.76$ & $100.00$ & $13.51$ \\
V-JEPA~2.1 attentive probe                & $63.51$ & $63.51$ & $81.08$  & $45.95$ \\
\ours{} (\retrtraj{})                     & $81.08$ & $81.08$ & $70.27$  & $91.89$ \\
\midrule
\ours{} (\oracletraj{})                   & $\mathbf{93.24}$ & $\mathbf{93.24}$ & $\mathbf{91.89}$ & $\mathbf{94.59}$ \\
\bottomrule
\end{tabular}
\end{table}

\paragraph{Findings.}
\emph{1) Without trajectory the task collapses.} \notraj{} predicts
``made'' on $69$ of $74$ test shots (all $37$ made and $32$ of $37$
missed), reaching only chance Balanced Acc. The pixel-only
V-JEPA~2.1 probe with a more expressive attentive head does better
but still leaves a $\sim\!18$-pp Balanced Acc.\ gap to the
trajectory variants.
\emph{2) Retrieved trajectory closes most of the gap.} \retrtraj{}
reaches $81.1\%$ Balanced Acc., $+24$ pp over \notraj{} and $+18$ pp
over the visual-only probe, against a $93.2\%$ Oracle upper bound.
\emph{3) Trajectory carries most of the outcome signal.}
The Oracle-vs-\retrtraj{} gap ($+12$ pp) is the room left for better
trajectory selection at test time, the same selection bottleneck
identified in \S\ref{sec:exp:atomic}. The split is small
($n_{\text{test}}{=}74$); margins on the order of $10$+ pp are
statistically meaningful, but individual cell values should be read
with that sample size in mind.

\section{Qualitative results}
\label{app:qualitative_results}

\begin{figure}[t]
\centering
\includegraphics[
width=0.92\linewidth,
trim=0 200pt 0 0,
clip
]{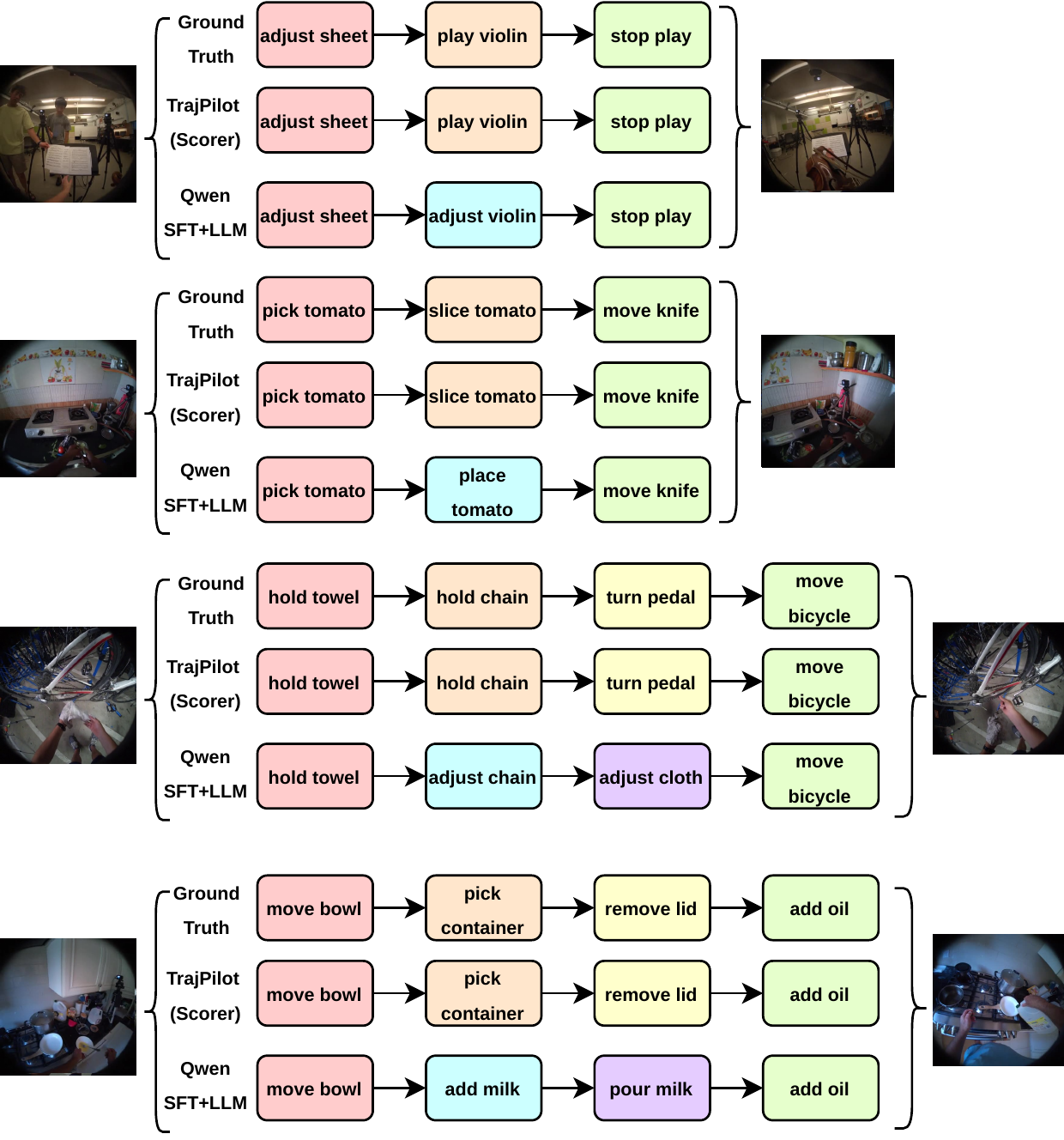}
\vspace{-0.5em}
\caption{\textbf{Qualitative open-vocabulary planning examples.}
Each block shows the observed start clip on the left, the observed
goal clip on the right, and the predicted action sequences. \ours{}
(\retrtraj) matches the ground-truth action sequence in these
examples, while the Qwen-SFT$+$LLM baseline often predicts plausible
but incorrect intermediate actions (e.g., wrong object/action
substitutions or incorrect procedural steps).}
\label{fig:qualitative_open_vocab}
\end{figure}

We complement the quantitative open-vocabulary planning results with
representative examples from \egoexo{} atomic actions. Each example
shows the observed start clip on the left, the observed goal clip on
the right, and three action sequences between them: ground truth,
\ours{} (\retrtraj), and the Qwen-SFT$+$LLM baseline. The examples
are drawn from cases where \ours{} (\retrtraj) matches the
ground-truth action sequence exactly while Qwen-SFT$+$LLM does not.
Qwen-SFT$+$LLM often predicts plausible actions but makes object
substitutions or inserts locally reasonable yet procedurally wrong
intermediate steps, whereas \ours{} (\retrtraj) recovers the correct
sequence. Results in Figure~\ref{fig:qualitative_open_vocab}.

\section{Broader impacts}
\label{app:broader_impacts}

TrajPilot predicts near-future actions and outcomes from first-person
video. Positive applications include assistive guidance for skilled
procedural tasks (flagging deviations during cooking, assembly, or
medical procedures before errors compound), training and coaching
feedback in sports and crafts, and accessibility tools that anticipate
user intent from movement. Negative applications follow from the same
capability: because much of the predictive power comes from
head-mounted camera trajectory rather than scene content, the method
can infer intent and outcome from a relatively impoverished signal,
raising concerns about workplace surveillance and behavioral inference
where wearable cameras are deployed without informed consent. The
benchmarks used here (Ego-Exo4D, Ego4D, EPIC-Kitchens-100, EgoPER) are
released for research with documented consent processes; deployment
outside that setting should preserve those norms. We do not release a
foundation model or scraped dataset, and the model produces action
labels rather than pixels, limiting direct misuse pathways but not the
inferential ones above.

\clearpage

\end{document}